\documentclass[journal,twoside]{{IEEEtran}}
\usepackage{amsfonts}
\usepackage{amssymb}
\usepackage{graphicx}
\usepackage{subcaption}
\usepackage{color}
\usepackage{multirow}
\usepackage[ruled,vlined]{algorithm2e}
\usepackage{tikz}
\usepackage{pgfplots}
\usepackage{pgfplotstable}
\usepackage{pgf-pie}
\usepackage{wrapfig}
\usepackage{url}
\usetikzlibrary{external}
\usetikzlibrary{timeline,chains}
\usetikzlibrary{mindmap,trees}
\usetikzlibrary{decorations.pathreplacing,positioning, arrows.meta}

\tikzset{
legend to the south/.code={
\coordinate[xshift=-1.5cm,
yshift=(\value{pgfpie@sliceLength}*0.5+1)*0.5cm-4em] (legendpos) at
(current bounding box.south);},
}
\begin{document}

\title{Fraud Review Detection:  Methods, Challenges, and Analysis}

\author{
Saeedreza Shehnepoor*,
% \orcidID{0000-0001-6760-4501} \and
Roberto Togneri,
% \orcidID{0000-0002-3778-4633} \and
Wei Liu,
% \orcidID{0000-0002-7409-0948} \and
Mohammed Bennamoun,
% \orcidID{0000-0002-6603-3257} 
\thanks{S. Shehnepoor (*corresponding author), 
R. Togneri,
M. Bennamoun, and
W. Liu is with the University of Western Australia, Perth, Australia.
emails: \{saeedreza.shehnepoor@research.uwa.edu.au, roberto.togneri@uwa.edu.au, wei.liu@uwa.edu.au, mohammed.bennamoun@uwa.edu.au.\}}
}
\maketitle

\begin{abstract}
Social reviews have dominated the web and become a plausible source of product information. 
% People and businesses use such information for decision-making. 
To gain more customers, businesses can hire 
% also make use of social information to spread fake information using 
a single user, groups of users, or a bot trained to generate fraudulent content. 
% Many studies proposed approaches based on user behaviors and review text to address the challenges of fraud detection. 
To provide an exhaustive survey on fraud review detection, we propose a framework that categorises the large volume of works in this area using three key components: the review itself, the user who carries out the review, and the item being reviewed. 
The literature is reviewed based on behavioral, text-based features and their combinations. With this framework, a comprehensive overview of approaches is presented including supervised, semi-supervised, and unsupervised learning. The supervised approaches for fraud detection are introduced and categorized into two sub-categories; classical, and deep learning. The lack of labeled datasets is identified with potential solutions suggested. To assist new researchers in understanding the field, a summary of future directions is included at each stage of the proposed systematic framework.
\end{abstract}



\keywords{ Fraud Review, Components, Features, Classification, Deep Learning}


\section{Introduction}
\label{sec:intro}

% social review platform has created a whole new type of relationships using internet infrastructure, where every individual could create a relationship in terms of friendship, collaboration, supervision, etc. At the early stages of social review platform, the users' activities were limited to a number of pre-determined tasks such as sharing media (text, images, videos), surfing the web. 
User generated data in the form of social reviews is one of the defining features of Web 2.0. This has motivated businesses to not only understand, but also take advantage of the opportunity presented by social reviews to promote their products and services, resulting in their own benefit and potentially the loss of their competitors. 
% This has compelled businesses to not only understand but also seize the opportunity provided by social reviews to promote services and products, 
% through advertisement and to provide an opportunity for customers to provide reviews, not only to other consumers benefits but also for businesses to improve their services and products. 
% Such an opportunity could set up a competitive situation where businesses try to improve their services to gain consumers' attention (reflected in their comments and reviews)
% resulting in their own gain and potentially competitors' loss. 
% It is worth to mention that every individual is able to create different identities in social review platform, which 
This unfortunately also motivates some businesses to employ fraudsters to spread fraudulent information through social review platforms such as Amazon and Yelp. 
\subsection{Definitions}
\label{sec:defs}
To facilitate understanding, we will first define the following key concepts:
\begin{itemize}
    \item \textbf{Fraud Review:} A review intentionally written by fraudsters to promote/demote services to boost gain/loss of the businesses providing the services. 
    \item \textbf{Fraudster:} A user paid by a business to write fraud positive reviews on their own business to promote services and products or to write fraud negative reviews on their competitors' services and products. 
    \item \textbf{Fraudster Group:} Fraudsters may form a group to collectively promote (demote) their own (competitors') services to dominate the sentiment of a service.
    \item \textbf{Target Item:} The service or product that a fraudulent review is written about.
    % The service or the product targeted by a fraud review is written for.
\end{itemize}
% One way to propagate fake information is to target a business through writing fraud reviews~\cite{Shebuit2015}. 
\begin{figure*}
    \centering
    \includegraphics[width= 19cm]{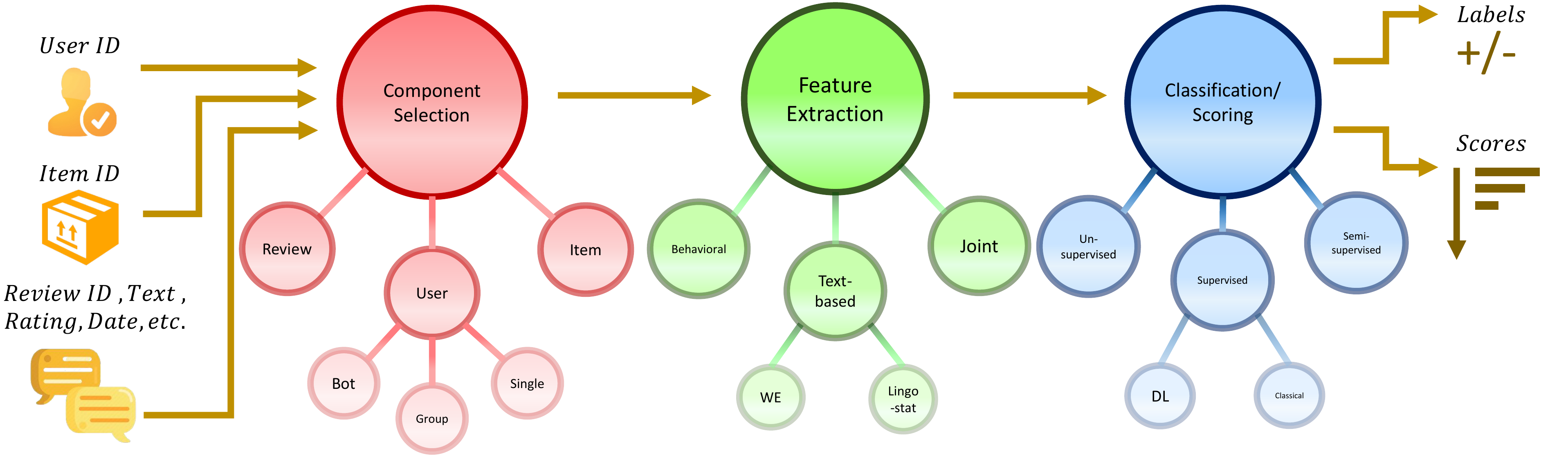}
    \caption{A systematic framework of steps in a typical fraud detector.}
    \label{fig:framework}
\end{figure*}
\subsection{Fraud Review in Comparison with Other Types of Spam}
\label{sec:spam-comparison}
Compared with other spam contents (e.g., email spams, insults, threats, malicious links, and fake news), fraud review detection is more challenging.

\textbf{Insults and threat} detection can rely on \textit{sentiment analysis} to find abusive comments~\cite{chiramel2020detection,van2018automatic,salminen2020developing}. \textbf{Malicious links} are detectable through blacklists~\cite{choi2011detecting,mamun2016detecting,agbefu2013domain,rashkin2017truth}. Building blacklists are among the most popular approaches to block malicious links. The blacklists are provided through several websites such as jwSpamSpy\footnote{\url{http://www.jwspamspy.net}}, PhishTank\footnote{\url{http://www.phishtank.com/}}, and DNS-BH\footnote{\url{http://www.malwaredomains.com}}. Such blacklists are built using user feedbacks and mechanisms to detect malicious URLs.

\textbf{Fake news} refers to articles intentionally written to convey false information for financial and political purposes. Though knowledge of political science, journalism, and psychology is needed to detect fake news, such contents are still detectable through fact-checking techniques as factual truths are available~\cite{shu2017fake,rashkin2017truth,pathak2019breaking}. 

\textbf{Fraud reviews}, on the other hand, are written or automatically generated in a similar way to genuine reviews. Hence, it is reported that even expert human judges find it hard to ascertain whether a review is fraud or not~\cite{Ott2011}.
Therefore, characterizing and detecting fraud reviews is not as simple as other types of spams. Additionally, studies~\cite{Zervas2016} show that fraud reviews increased in Yelp by 5\%
%\footnote{\href{https://www.businessinsider.com.au/20-percent-of-yelp-reviews-fake-2013-9?r=US&IR=T}{A Whopping 20\% Of Yelp Reviews Are Fake}} 
to 25\% from 2005-2016, while a rating increase of 1-star in Yelp may lead to a 5-9\% increase in revenue for a restaurant~\cite{YelpRate2018}. Hence, fraud detection is important to social review services such as Yelp and Amazon to provide spam-free platforms for users. As illustrated in Fig.~\ref{fig:framework}, a social review detector can be viewed as the interaction of three types of constituents, namely, User, Review and Item. Computationally, a social review detector can be represented as a set of triplets $\langle$User, Review, Item$\rangle$, which denotes the fact that a \textit{user} wrote a \textit{review} about an \textit{item}. To facilitate using a different annotation, their description is given in Table~\ref{tab:not-list}.

\begin{center}
\begin{table}[hbt!]
% \captionsetup{font=8p}
\centering
\caption{List of symbols and their corresponding descriptions.}\label{tab:not-list}
\begin{tabular}{|c|p{7cm}|}
\hline
Symbol &  Description \\
\hline
\hline
$R_u$ & Review written by user $u$\\ \hline
$X^i$ & $i^{th}$ element of component $X$ (e.g. $R^{i}_{u}$ is $i^{th}$ review of user $u$)\\ \hline
$N_{X_u}$ & Number of $X$ attribute on user $u$\\ \hline
$NR_u$ & Negative review written by user $u$\\ \hline
$PR_u$ & Positive review written by user $u$\\ \hline
$Rating_u$ & Rating given by user $u$\\ \hline
$avg(X_u)$ & Average of $X$ attribute on user $u$\\ \hline
$R_{u-p}$ & Review written from user $u$ on item $p$\\
\hline
\end{tabular}
\end{table}
\end{center}

% In the past decade, a number of review articles have attempted to provide comprehensive overviews of fraud reviews detection methods, as described below. 

% , different review papers have investigated the area. 
% Heydari \textit{et al.}~\cite{heydari2015detection} performed a review on fraud review and fraudster detection. 
% primarily studsy spam detection from a data point of view. This study 
% Heydari \textit{et al.}

\subsection{Fraud Review Categorization and Identification - Past Literature Surveys}
\label{sec:fraud-id}
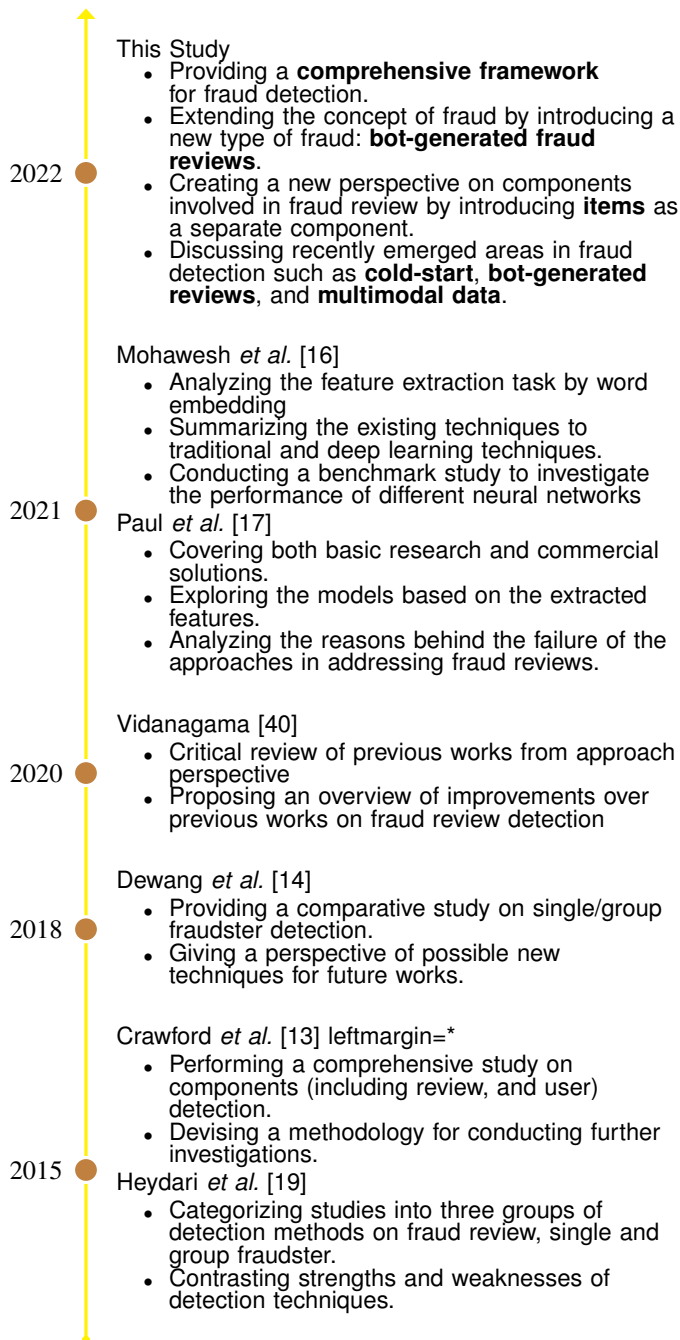
\begin{figure}
\begin{center}
            \begin{tikzpicture}[
node distance = 3mm and 5mm,
  start chain = A going below,
  dot/.style = {circle, draw=white, very thick, fill=brown,
                 minimum size=3mm},
  box/.style = {rectangle, text width=75mm,
                 inner xsep=4mm, inner ysep=1mm,
                 font=\sffamily\small\linespread{0.84}\selectfont,
                 on chain},
                        ]
    \begin{scope}[every node/.append style={box}]
\node {This Study
\begin{itemize}
    \item Providing a \textbf{comprehensive framework} \\for fraud detection. 
    \item Extending the concept of fraud by introducing a new type of fraud: \textbf{bot-generated fraud reviews}.
    \item Creating a new perspective on components involved in fraud review by introducing \textbf{items} as a separate component. 
    % \item Considering recently introduced techniques by \textbf{DL} on both \textbf{feature extraction (WE)} and \textbf{classification} level. 
    \item Discussing recently emerged areas in fraud detection such as \textbf{cold-start}, \textbf{bot-generated reviews}, and \textbf{multimodal data}.
\end{itemize}
} ;
\node {Mohawesh \textit{et al.}~\cite{mohawesh2021fake}
\begin{itemize}
    \item Analyzing the feature extraction task by word embedding 
    \item Summarizing the existing techniques to traditional and deep learning techniques.
    \item Conducting a benchmark study to investigate the performance of different neural networks  
\end{itemize}
Paul \textit{et al.}~\cite{paul2021fake}
\begin{itemize}
    \item Covering both basic research and commercial solutions.  
    \item Exploring the models based on the extracted features.
    \item Analyzing the reasons behind the failure of the approaches in addressing fraud reviews. 
\end{itemize}
} ;
\node {Vidanagama~\cite{vidanagama2020deceptive}
\begin{itemize}
    \item Critical review of previous works from approach perspective
    \item Proposing an overview of improvements over previous works on fraud review detection
\end{itemize}
} ;
\node {Dewang \textit{et al.}~\cite{dewang2018state}   
\begin{itemize}
    \item Providing a comparative study on single/group fraudster detection.
    \item Giving a perspective of possible new techniques for future works.
\end{itemize}
} ;
\node {Crawford \textit{et al.}~\cite{crawford2015survey}
\begin{itemize}[leftmargin=*]
    \item Performing a comprehensive study on components (including review, and user) detection. 
    \item Devising a methodology for conducting further investigations.
\end{itemize}
Heydari \textit{et al.}~\cite{heydari2015detection}
\begin{itemize}
    \item Categorizing studies into three groups of detection methods on fraud review, single and group fraudster.
    \item Contrasting strengths and weaknesses of detection techniques.
\end{itemize}
} ;
    \end{scope}
\draw[very thick, yellow, {Triangle[length=4pt]}-{Circle[length=3pt]},
      shorten <=-3mm, shorten >=-3mm]           % <--- here is adjusted additional arrow's 
    (A-1.north west) -- (A-5.south west);
\foreach \i [ count=\j] in {2022, 2021,2020,2018,2015}
    \node[dot,label=left:\i] at (A-\j.west) {};
    \end{tikzpicture}    
    % \end{minipage}
   
\end{center}
% \begin{minipage}{\textwidth}
 
% \begin{minipage}{\textwidth}
%     \begin{tikzpicture}
%         \centering
%         \includegraphics[height=13cm,right]{samples/Figs/Survey-fig-v2.pdf}
% % \caption{Timeline of recent review papers.}
% % \label{fig:componentsTime}
%     \end{tikzpicture}
% \end{minipage}
    
    \caption{The timeline of recent review papers.}
    \label{fig:time-line}
\end{figure}

In the research community, recent studies focused on the identification of fraud reviews and their characteristics. Fig. \ref{fig:time-line} summarizes the main topics covered by different fraud detection review papers. 
\subsubsection{Fraud Definitions and Categorization}
\label{sec:fraud-cat}
Crawford \textit{et al.}~\cite{crawford2015survey}, and Dewang \textit{et al.}~ \cite{dewang2018state} adopted an analytical approach to identify fraud reviews. Dewang \textit{et al.}~\cite{dewang2018state} considered fraud reviews as one of the many spam types including email spams, web spams (which refers to techniques used to manipulate the rank and the orientation of web pages), and blog spams (defined as messages posted randomly on different discussion boards), while Crawford \textit{et al.}~\cite{crawford2015survey} survey different types of fraud reviews, using the definitions provided by Jindal \textit{et al.}~\cite{Jindal2008}:
% same as the definitions given by~\citet{Jindal2008}:
\begin{itemize}
    \item \textbf{Untruthful Reviews:} Such reviews are written to promote/defame products intentionally. 
    \item \textbf{Fan Reviews:} The second type is reviews written by extreme/radical fans of specific brands or products. 
    % The second type is the reviews written for particular brands or products by radical fans.
    \item \textbf{Junk Reviews:} These reviews contain no relevant information regarding the targeted item; e.g., advertisements.
\end{itemize}
Crawford \textit{et al.}~\cite{crawford2015survey} then focus on  the detection of untruthful reviews, similar to the current study.  Mohawesh \textit{et al.}~\cite{mohawesh2021fake} proposed a similar categorization to that of Jindal \textit{et al.}~\cite{Jindal2008} and provided two examples to explain the challenging task of labeling the reviews by human. To provide a more general category for fraud detection, Paul \textit{et al.}~\cite{paul2021fake} categorized fraud reviews as a type of opinion spam, alongside other categories such as advertisements, unrelated random texts, and product questions. 

Given all definitions and different types of fraud reviews investigated by various review papers, some new types of fraud reviews are still missing. Bot generated reviews, are still missing as an important type of fraud review which are particularly difficult to spot~\cite{Yuanshun2017}. In this paper we also consider bot generated reviews as another type of fraud and investigate relevant recent studies. 

\subsubsection{Review Data}
Heydari \textit{et al.}~\cite{heydari2015detection} categorized review data into two major subcategories: \textbf{1)} the content of reviews and \textbf{2)} the metadata of reviews.
The content of the reviews is used to extract text-based features, and the metadata is employed to extract behavioral features (to be discussed in Sec.~\ref{sec:cat-features}). Most datasets provide both review text and metadata (e.g., rating, date of the review, user ID, and item ID) alongside the ground-truth. The statistics of major datasets is provided in Table \ref{tab:datasets}.
\begin{table*} 
\centering
  \caption{
Details of datasets in previous studies.}
  \label{tab:datasets}
  \begin{tabular}{ccccm{5cm}m{3cm}}
    \hline
    Datasets & Reviews & Users & Items & Metadata & Ground-truth\\ \hline
    Yelp & 608,598 & 260,277 & 5,044 & review text, rating (1-5), date, user ID, item ID & Yes (Yelp recommender system)~\cite{Shebuit2015} \\
    Amazon & 53,777 & 42,655 & 6,822 & review text, rating (1-5), date, user ID, item ID, profile name, feedback (number of likes) & Yes (human judges)~\cite{Jindal2008} \\
    TripAdviser & 1600 & - & 20 & review text, rating (like, dislike), date, user ID, item ID & Yes (human judges)~\cite{Ott2011} \\ 
    Tencent & 302,097 & 82,542 & 7,584 & review text, rating (1-5), date, user ID, item ID & Yes (near ground-truth)~\cite{Wang2019FdGars} \\ 
    ResellerRating & 408,470 & 343,063 & 14,561 & rating (1-5), user ID, item ID & Yes (human judges)~\cite{wang2011review} \\
    MT & 2,836 & - & - & review text, rating (1-5), user ID, item ID & Yes (Amazon Mechanical Turk)~\cite{crawford2016reducing}\\ 
    Dianping & 21,255 & 504 & 14,187 & review text, rating (1-5), date, user ID, item ID & Yes (human judges)~\cite{Li2016}\\ 
    Epinions & - & 189,028  & - & rating (1-5), trust, user ID & No~\cite{Yoo2017}\\ 
    WeiDai & 36,851 & -  & - & review text, user ID & Yes (recommender)~\cite{He2020}\\ 
    Lending & 18,405 & -  & - & user ID & Yes (recommender)~\cite{Zhao2019}\\ 
    TREC & 707,664 & -  & - & review text, user ID & Yes (TREC spam tracker)~\cite{Lai2011}\\ 
    Expedia & 1,800 & -  & - & review text, review rating, user ID, item ID & Yes (human judge)~\cite{Banerjee2015}\\ 
    Google & 341,993 & 265,724  & 718 & date, rating (1-5), user ID, item ID & Yes (human fraud workers)~\cite{Hernandez2018}\\ 
    Priceline & 4,427 & -  & 118 & date, review text, user ID, item ID & Yes (Amazon MT)~\cite{Myle2012}\\ 
    Weibo & 9,726 & -  & - & date, review text, user ID, item ID & Yes (human judge)~\cite{Wang2016}\\ 
    JD & 48,562 & -  & 100 & date, rating (like, dislike), user ID, item ID, uselessVoteCount, isMobile & Yes (human judge)~\cite{Deng2017}\\ 
    SWM & 1,132,373 & 966,842 & 15,094 & date, rating (1-5), user ID, item ID & No~\cite{Akoglu2013}\\ 
    \hline
    \end{tabular}
\end{table*}
Mohawesh \textit{et al.}~\cite{mohawesh2021fake}, however, proposed a different categorization where metadata is considered as a subcategory of the text-based features. The latest review by Paul \textit{et al.}~\cite{paul2021fake} surveyed models based on the features extracted by different approaches. Their features were extracted from both the review texts and metadata.

Although previous survey papers provided a thorough review on the data that could be utilized for fraud review detection, they do not consider new modalities of data (IP, MAC address, etc.) that could be employed for fraud detection. This multi-modal data has been utilized before for untruthful reviews in Twitter~\cite{liu2019opinion}. We will discuss the potential of such data in providing a better overview of the user activity on a platform in Sec.~\ref{sec:discussions}. The multimodal dataset helps the fraud detection algorithm to deal with more complicated tasks such as the cold-start problem~\cite{wang2017handling}. The ``cold-start problem" for fraud detection refers to the challenge of detecting fraud when there is little or no previous data available to use for training a model. This can make it difficult to accurately identify fraudulent activity.

Few recent studies~\cite{maurya2022deceptive,mewada2022comprehensive} also focused on components involved in fraud review detection, failing in delivering a systematic framework used in fraud review detection.   
\subsubsection{Components of Interest}
Heydari \textit{et al.}~\cite{heydari2015detection}, Dewang \textit{et al.}~\cite{dewang2018state}, and Crawford \textit{et al.}~\cite{crawford2015survey} also categorized the fraudster detection tasks into different categories. Heydari \textit{et al.}~\cite{heydari2015detection} and Dewang \textit{et al.}~\cite{dewang2018state} considered two categories of single and group fraudster detection, while Crawford \textit{et al.} \cite{crawford2015survey} mostly focused on two different tasks of \textbf{1)} review-based fraud detection and \textbf{2)} user-based fraud detection. 

None of the current literature surveys examined the importance of targeted items (products or services) in identifying the fraud contents and preventing them to spread. In contrast to previous surveys, we consider targeted items as another component of interest and examine their potential to address the recent challenges on fraud detection, as utilized by Ji \textit{et al.} ~\cite{ji2020burst} for the cold-start problem. 
\subsubsection{Features and Techniques}
\label{sec:cat-features}
Heydari \textit{et al.}~\cite{heydari2015detection} and Crawford \textit{et al.}~\cite{crawford2015survey} elaborated on different features for the detection task. Dewang \textit{et al.}~\cite{dewang2018state} specifically divided the features into linguistic and non-linguistic features. Heydari \textit{et al.}~\cite{heydari2015detection} then divided fraud review detection techniques into two major groups: duplicating detection methods and content-based methods. Crawford \textit{et al.}~\cite{crawford2015survey}, Dewang \textit{et al.}~\cite{dewang2018state}, and Vidanagama \textit{et al.}~\cite{vidanagama2020deceptive} divided the classification approaches into supervised, unsupervised, and semi-supervised. In contrast to previous surveys, Vidanagama \textit{et al.}~\cite{vidanagama2020deceptive} provided a critical review on features and techniques through comparison and also considered network-based approaches, overlooked by previous studies.  Mohawesh \textit{et al.}~\cite{mohawesh2021fake} surveyed the features with more focus on the modality of extracted features. Mohawesh \textit{et al.}~\cite{mohawesh2021fake} were the first to survey word embedding techniques and explored different studies to cover the fraud detection approaches that employed such techniques. Paul \textit{et al.}~\cite{paul2021fake} specifically focused on the models based on the extracted features. Paul \textit{et al.}~\cite{paul2021fake}, therefore, categorized Fake Review Detection (FRD) models into those which employed Content Similarity (FRD-CS), Writing Footprints (FRD-WF), Behavioral Footprints (FRD-BF), Network Footprints (FRD-NF), Rating Footprints (FRD-RF), and finally Collusion Footprints (FRD-CF). 

Previous review papers considered different semi/unsupervised learning approaches to show their importance in dealing with fraud review detection when there is a lack of data. However, many recent studies utilized graph representation learning to show their potential in addressing such a challenge, such as Graph Convolutional Network~\cite{Zhang2020gcn}, graph-based inductive learning~\cite{Hooi:2017:GFD:3119906.3056563}, etc. We will discuss such techniques in Sec.~\ref{sec:semi-supervised} and Sec.~\ref{sec:unsupervised} to provide a better overview of such techniques. 

\subsubsection{Summary}
Fraud detection is a relatively new and evolving area, some challenges are only partially addressed while new challenges have recently emerged. For example, there are also long-standing challenges that are still relevant (e.g. lack of data), and new challenges to tackle (bot-generated reviews, cold-start problem, etc.). Exploring all these topics under a unified framework helps researchers to acquire new insights of future directions in fraud review detection.

\subsection{Scope of This Review}
Although previous studies have contributed to a better understanding of fraudulent reviews and fraudsters, there is currently no comprehensive framework that covers all aspects of fraud detection. The framework should be simple and easy to understand while helping define the scope of different studies and to clarify future challenges.
% The systematic framework showing the steps of a typical fraud detection framework is depicted in Fig.~\ref{fig:framework}. To fully characterize a fraud detection framework, three steps are required. 
Hereby, we devised a framework based on three essential steps in fraud detection tasks, as illustrated in Fig.~\ref{fig:framework}.
\subsubsection{Component Selection}
In the first step, a component should be selected for information collection. The framework includes three different components: review, user, and item. Review, single-user, and group users are mentioned in previous review papers. Previous review papers only considered the fraud and fraudster (single, group) detection task, but not targeted item identification. The targeted items recently attracted a lot of researchers' attention~\cite{kumar2018rev2,Shebuit2015}, and were detected in combination with fraud review and fraudsters. Fig.~\ref{fig:framework} depicts the different components which can be selected by a fraud review detection algorithm. This review filled the gap by surveying works on targeted item detection.
% Several studies proposed fraudster detection approaches based on the user behavior analysis \cite{Ee-Peng2010,kumar2018rev2,Akoglu2013}. On the other hand, fraudsters use different strategies to attack a business. Such strategies are studied by multiple papers through finding single fraudsters \cite{Jindal2008} in early stages, while with formation of fraudster group higher destructive impacts were observed on businesses \cite{Arjun2012,Akoglu2013,Wang2012,shehnepoor2021hinrnn}.\\
Another overlooked emerging area of study includes bot review generation. Such generators cause serious problems in social review platforms. Recent studies on fraudsters indicated that they are now using bot-generated reviews. Bots simply generate reviews similar to human written reviews, and simultaneously bots hide their footprints \cite{Yuanshun2017}.

% In this study, we also investigate targeted item detection, as a topic overlooked by previous studies.  
% \begin{figure}[h]
%   \centering
%     \begin{tikzpicture}[scale = 0.8]
%       \path[mindmap,concept color=green!40!black,text=black]
%         node[concept ] {Feature}
%         [clockwise from=-30] 
%         child[concept color=green] {
%           node[concept] {Joint}
%         }
%         child[concept color=green] { node[concept] {Text-based} 
%         [clockwise from=-60]
%           child[concept color=green!30!white] { node[concept] {Statistical} }
%           child[concept color=green!30!white] { node[concept] {Vector Representation} }
%         }
%         child[concept color=green] {
%           node[concept] {Behavioral}
%          };
%     \end{tikzpicture}  
%   \caption{Structural chart of previous features covered in previous fraud detection surveys.}
%     \label{fig:features}
% \end{figure}
%we also focus both on fraud detection and a new area of sophisticated fraud generation as an emerging problem. 
\subsubsection{Feature Extraction}
The second step extracts features from and for the selected component.

Text-based features refer to features that are directly extracted from the text using language models~\cite{Song2012} or text statistics~\cite{Chang2015,Li2011,Myle2012}. To improve the review and user representations, review metadata is used to extract behavioral features~\cite{Jindal2008,J.2015,Arjun2012}. Text-based and behavioral features are combined as ``joint" features to achieve a better representation. Subsequent studies also used the same joint representation to achieve better components' representation~\cite{Shehnepoor2017,Shebuit2015}. Given the fair performance of the lingo-statistic features and their combination with other types of features they can still be manipulated by fraudsters to escape detection. Recent advances in neural language models, such as  Word Embeddings (WEs) can be used to produce a vector representation of review texts~\cite{Aghakhani2018,Yafeng2016,Duyu2015,Quoc2014} to overcome the limitations of text-based lingo-statistic features in achieving a global representation. In other words, lingo-statistic features suffer from a limitation to extract the sentiments from different aspects of a text, while WE is capable of extracting sentiment from various aspects of a given text. Fig.~\ref{fig:framework} provides an overview of some of the utilized features in fraud detection. We extend on the previous reviews by covering a wider range of review text representation learning.
% \begin{figure}[h]
%   \centering
%     \begin{tikzpicture}[scale = 0.8]
%       \path[mindmap,concept color=blue!40!black,text=white]
%         node[concept ] {Approach}
%         [clockwise from=-30]
%         child[concept color=blue] {
%           node[concept] {Semi-supervised}
%         }
%         child[concept color=blue] { node[concept] {Supervised} 
%         [clockwise from=-60]
%           child[concept color=blue!60!white] { node[concept] {classical} }
%           child[concept color=blue!60!white] { node[concept] {Deep Learning} }
%         }
%         child[concept color=blue] {
%           node[concept] {Un\-supervised}
%          };
%     \end{tikzpicture}  
%   \caption{Structural chart of the previous approaches which were covered in previous fraud detection surveys.}
%     \label{fig:approaches}
% \end{figure}

\subsubsection{Classification/Scoring}
In the third and final step, a classifier or scorer is applied to the extracted features for the final labeling (scoring). Such approaches are categorized as unsupervised, semi-supervised, or supervised. Due to the scarcity of trusted labeled data in fraud detection, recent studies employed unsupervised learning \cite{Hooi:2017:GFD:3119906.3056563,Shebuit2015} and semi-supervised \cite{Shehnepoor2017} approaches. Supervised learning is mostly applied on the Yelp dataset \cite{Aghakhani2018,Ott2011a} for fraud detection. Fig.~\ref{fig:framework} displays an overview of fraud detection approaches. Deep learning is the most recent popular supervised learning approach for improving fraud detection accuracy~\cite{Yafeng2016,Shaohua2018,Aghakhani2018}.\\ 
To highlight the novelties of this survey, Fig.~\ref{fig:framework} presents each step of the framework with a focus on the overlooked aspects in previous review papers.
In summary, this survey paper offers the following contributions:
\begin{itemize}
    % \item Providing a comprehensive framework for \textbf{fraud detection}. 
    % \item Extending the concept of fraud by introducing new type of fraud: \textbf{bot generated fraud reviews}
    % \item Creating a new perspective on components involved in fraud review by introducing \textbf{items} as a separate component 
    % \item Considering recent introduced techniques by \textbf{DL} on both \textbf{feature extraction (VR)} and \textbf{classification} level. 
    % \item Discussing recently emerged areas in fraud detection such as \textbf{cold-start}, \textbf{bot generated reviews}, and \textbf{mutlimodal data}
    \item
    %We create a new perspective on components involved in fraud detection. 
    For the first time, we introduce bot fraud generation as a newly emerging area in fraudulent reviews. 
    \item We also extend the components to items targeted by fraudsters and explore the studies in the target item detection. 
    \item 
    We provide a more complete overview of open problems and identify three categories of classical, ongoing, and hot topics to facilitate better understanding of future directions of the fraud review detection research. 
    % each step of our proposed framework and discuss how each classified component (in combination with graph-based approaches) can facilitate the fraud detection task. 
    In each category, we then explain the usefulness of such approaches to address future challenges, such as cold-start \cite{wang2017handling,you2018attribute,dfraud}, bot-generated reviews \cite{Hooi:2017:GFD:3119906.3056563}, transfer learning, fraud detection through multimodal data~\cite{liu2019opinion}, end etc. 
    \item 
    We provide the first comprehensive survey that covers various research by following a systematic framework for fraud detection in social review platforms.
    % in a review paper. The framework (Fig.~\ref{fig:framework}) introduces three main steps for fraud detection which will be elaborated in their respective sections.
    % , \ref{fig:fields}, \ref{fig:features}, and \ref{fig:approaches}. 
    % in each fraud detection system, with each step given the main topics in the corresponding area.
    % \item
    % %We study the recent techniques employing deep learning for feature extraction and classification, as one of the promising topics in fraud detection. 
    % We discuss the deep learning approaches for the first time and explain how such techniques accelerate the detection of fraud in social review platforms for both feature extraction and classification/scoring. 
% \item A thorough analysis of the features used in fraud detection, including text-based, behavioral, and the joint representation of text-based and behavioral is discussed. We also review the studies on the non-handcrafted features such as vector representation (VR) features, which were overlooked in previous reviews. 
    % \item 
    % We provide a comprehensive analysis of the classical and deep learning classifiers that are used in fraud detection.     \item
    % Finally, upcoming areas and future challenges are discussed to provide a better overview of the possible directions for fraud detection. 
\end{itemize}
\begin{figure}
    \includegraphics[width=\linewidth]{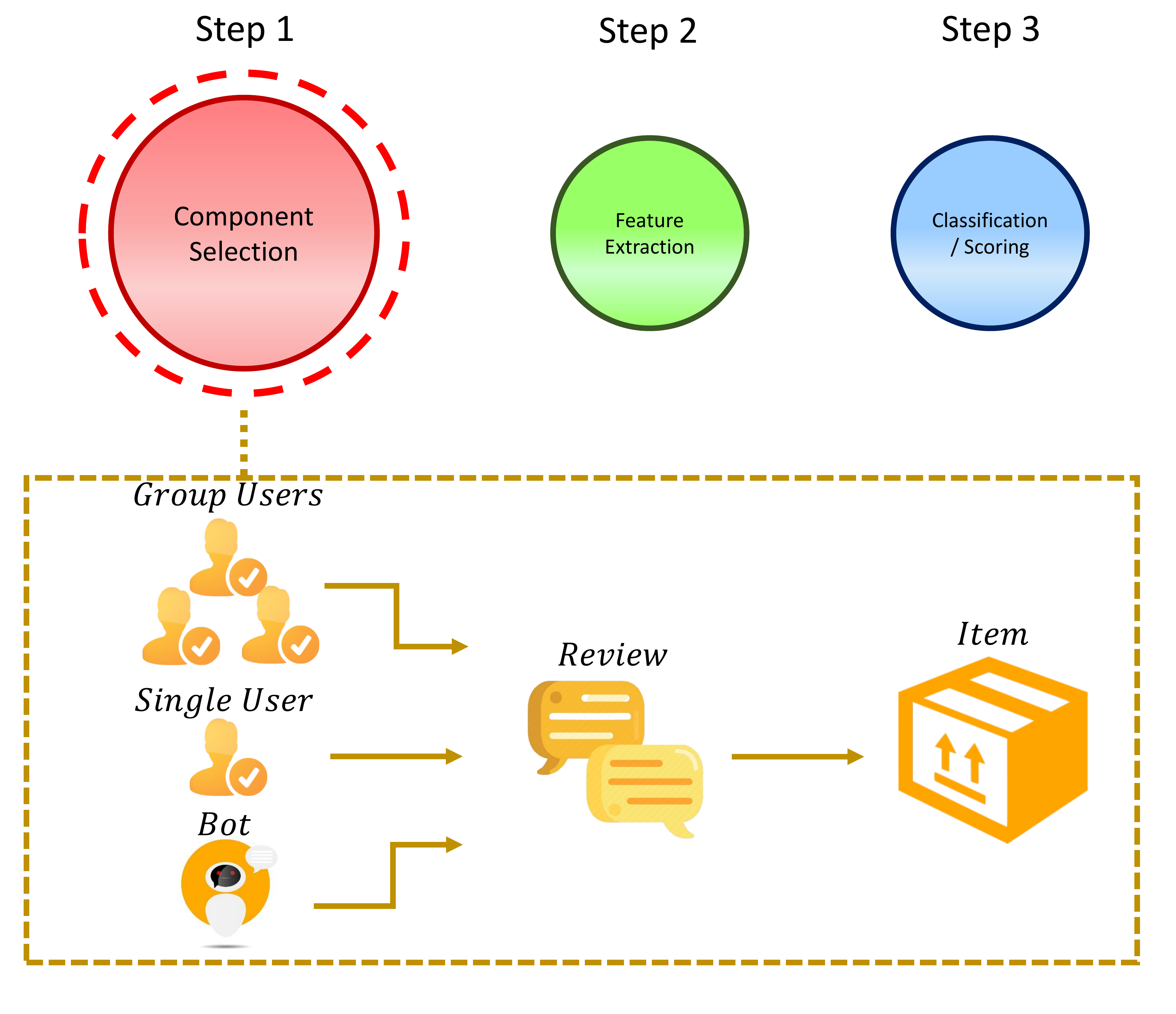}
    \caption{An overview of component selection step in a fraud detection framework.}
    \label{fig:comp-step}
\end{figure}  

In the following sections, we first review the fraud detection studies focusing on each of the three steps of our proposed fraud detection framework, including components in Sec.~\ref{sec:component}, features in Sec.~\ref{sec:features}, and approaches in Sec.~\ref{sec:approaches}, respectively. Next we provide discussions on the datasets, topic analysis in fraud detection, and future directions based on the new identified challenges, in Sec.~\ref{sec:discussions}. Finally we conclude our survey in Sec.~\ref{sec:conclusion}. 

\section{First Step: Component Selection}
\label{sec:component}

% The first step of the fraud detection task is to select the component. 
Since the first review work by Jindal \textit{et al.}~\cite{Jindal2008} in 2008, the scope of most studies are limited to fraud review and fraudster detection. Here, we broaden the scope by introducing targeted item detection while expand the concept of fraudsters to include software bots. Bots are capable of generating contents, in a fashion that are highly similar to those written by human. 
% Additionally, we also explore the studies on targeted item detection.
Fig. \ref{fig:comp-step}. shows the three key components involved in fraud review detection. In the following, we discuss specific studies which considered a single component as their main focus. 
% We also present the studies focused on features and classification approaches and discuss them in their respective sections. 
% Three involved components in fraud detection are discussed in this section; review, user, and item. Fraud review detection is first discussed. Then we present the latest studies on detection of different types of fraudsters, including bot users, involved in fraud review. 
% Bots are categorized as a separate fraudster and recent bot detection methods are covered. 
% Finally, targeted item detection is introduced and related studies are discussed.
% While in fraud detection ML approaches mostly detect the review attacks, recent novel approaches are proposed for detecting different components in social review platform. There are three different components in online social review platform; a user, a review written by a user, and an item targeted by a user. 
% In other words, the available data for a specific component is gathered for the next steps. 
% Hence, we divided this sections in three subsection, where in first section the problem of fraud review detection is investigated in a broad sense. In second section, the problem of fraud detection is introduced and extended for different types of users. Reviews are either written in by a single user or by a group of fraudsters. Furthermore, fraud detection and its recent introduced counterpart bot fraud generation are deeply correlated. To deal with this type of attack, new methods require to act effectively to reduce the effects of false information spread. Finally, the items are explored and discussed. 

\subsection{Review}
Early studies on fraud review detection mostly focused on fraud review detection as the main task. For the first time, Jindal \textit{et al.}~\cite{Jindal2008} proposed a categorization for different types of fraud reviews, namely, untruthful opinions, reviews on brands only, and non-reviews (as discussed in Sec.~\ref{sec:fraud-cat}). 
% Untruthful opinions (i.e. fraud reviews) refer to reviews written to mislead customers in their decision making. The reviews on brands are typically written on special brands by the fans. Non-reviews are written in specific patterns to maximize their impression (e.g. advertisements). 
Jindal \textit{et al.}~\cite{Jindal2008} employed different features such as review rating, review feedbacks, rating deviation, review ranking (behavioral), and review feedback (linguistics) to deal with fraud reviews. It is worth mentioning that Jindal \textit{et al.} split up features into three sets of categories; reviews, reviewer, and product-centric features. The results on the Amazon dataset showed an AUC of 98.7\% for fraud review detection. 

Jindal \textit{et al.}~\cite{Jindal2008} were the first to provide a list of employed features  for fraud review detection. Afterward, researchers proposed various methods to improve the performance of the fraud review detection.

Singleton fraud reviews are reviews written by the same person using different account names. Sandulescu \textit{et al.}~\cite{sandulescu2015detecting} was the first to  propose a framework to spot singleton reviews. The proposed framework specifically employed two different methods: first, the review similarity; second, the topic similarity using bag-of-words and bag-of-opinion-phrases. For topic modelling, the proposed framework utilized Latent Dirichlet Allocation (LDA). The proposed approach showed a maximum precision of 80\% on the Yelp dataset.

% Kauffmann \textit{et al.}~\cite{kauffmann2019framework} presented the concept of User-Generated Content (UGC) to conduct different analyses on reviews of the Amazon dataset. Kauffmann \textit{et al.} proposed a unified framework utilizing both review text and review rating for fraud review detection. To find the sentiment score of the reviews, Kauffmann \textit{et al.} performed three tasks of lexical analysis, synthetic analysis, and semantic analysis. Finally, the sentiment scores were combined with the review rating to calculate the probability of each review to be genuine or fraud. 
% calculated based on similarity measure  between each two reviews drawn from tf-idf metric. 
% Given different levels of semantic analysis, Kauffmann \textit{et al.} achieved an accuracy of 85.53\% on the Amazon dataset.\\
Barbado \textit{et al.}~\cite{barbado2019framework} proposed a framework that extracts several types of features for fraud review detection. The proposed features include personal features (e.g., a profile description, bookmarks, and updates), social features (e.g., popularity, and compliments), review activity (e.g., review count, negative ratio, and self-deviation), and trust (e.g., relative deviation, and content similarity). An explanation of a selected few of features is provided in Table~\ref{tab:features-bar}.

\begin{table}
\centering
\caption{An overview of some of the features employed by Barbado \textit{et al.}~\cite{barbado2019framework}. H/L depicts if a High/Low value of the
the feature is more likely to be associated with the fraud.}
\label{tab:features-bar}
\begin{tabular}{m{2cm}m{3cm}m{0.5cm}m{1.5cm}}
\hline
Features &  Explanation & H/L& Formula\\
\hline
\hline
Review Count & Number of reviews by a specific user & H & $N_{R_u}$\\ \hline
Negative Ratio & Ratio of negative review vs. total number of reviews & H & $\frac{N_{Rating_u=1,2}}{N_{R_u}}$\\ \hline
Self Deviation & Deviation of ratings given by the fraudster from the rating average value& H & $\frac{1}{N_{R_u}}\sum_{i}(Rating^i_u-avg(Rating_u))$\\ \hline

\end{tabular}
\end{table}

% The review centric features are typically text-wise features including sentiment analysis, tf-idf, and LDA. 
% Various classifiers are applied to the output of previous step and the best result is reported for
Barbado \textit{et al.} utilized an adaBoost classifier which yielded an F-measure of 82\% on the Yelp dataset. 
With the advance in context-aware applications, Ruan \textit{et al.}~\cite{ruan2020gadm} proposed a Geolocation-based Account Detection Model (GADM) to improve the fraud review detection performance. GADM utilized geolocation information alongside account information (e.g., review rating) to capture location-based information in three phases. In the first phase, account features and geolocation features were extracted from the reviews and used as input. The geolocation information was obtained from visited restaurants and hotels by users. The geolocation information is then fed to a Long-Short Term Memory (LSTM) to model the user location series. A classification step is then carried out on account information and the outputs were then combined and fed to SVM for final classification. GADM showed an accuracy of 85.8\% on the Yelp dataset. 
% Given the importance of features relevant in detecting fraud reviews (e.g. spatial information by Ruan \textit{et al.}), 

He \textit{et al.}~\cite{He2020} proposed a scheme considering the temporal relationship between reviews. He \textit{et al.}~\cite{He2020} proposed a PU learning based on positive (P) samples (fraud reviews) and unlabelled (U) ones. 
The PU learning refers to an approach, where unlabeled samples are scored based on the samples with positive labels, in a binary classification task. 
The proposed PU scheme includes three consecutive steps designed to score reviews being fraud or genuine. 

In the first step, the time interval between every two reviews was used as a feature to label unlabelled reviews as Real Negative (RN) reviews. 
% Different classifiers such as Naive Bayes (NB), Rocchio classifier, and 1-DNF are applied to extracted features. 
Next, an SVM classifier is trained using the Positive (P) and RN set. Finally, text-based features were combined with outputs of previous outcomes into an Expectation-Maximization (EM) algorithm for the final clustering. Data collected from the WeiDai-Financial\footnote{\url{http://weidai.investorroom.com/}} website were used as the main dataset. The proposed approach showed an F1-measure of 75.3\%. 

The analysis of studies on fraud review detection shows the importance of spatial and temporal modeling of reviews in a social review platform to detect fraud reviews. 

\subsection{User}
Initially, fraudsters worked in isolation when using a social review platform. As businesses recognised the significant influences of social reviews on sales and profits, fraudsters started working in groups to gain more impact. With the advances in Deep Learning techniques in recent years, bot fraudsters started to play a non-negligible role in generating and propagating fraud reviews. In this section, we take into account these new trends and focus on fraudster group and bot fraudster detection after a review of the traditional single fraudster detection. 
% introduce and present the studies on fraudster detection in three subsections: Single User, User Group, and Bots.

\subsubsection{Single User}
% As explained in Sec.~\ref{sec:defs}, fraudsters either work individually, or in a group to engage in fraudulent activities.
Single user refers to a fraudster who actively is involved in fraudulent activities as an individual. \textit{Single fraudster detection} techniques mostly rely on a users' behavior pattern and their actions. Jiang \textit{et al.} \cite{jiang2015general,jiang2016spotting} proposed a suspicious behavior formulation through a combination of matrix decomposition (for finding unexpected values) and KL Divergence for K-mode speciousness. A matrix is initialized first with each user as the row and features as the columns. For each user, the ID, IP, time of the written reviews and retweet comment were used to find fraud reviews on Twitter.

Chiu \textit{et al.}~\cite{chiu2017uncovering} also utilized user behavior analysis to find suspicious activity, using \textit{multimodal data}. \textit{Multimodal data} refers to data gathered from different sources such as security devices, networks, servers, and applications. Chiu \textit{et al.}~\cite{chiu2017uncovering} employed three modalities~\cite{baltrusaitis2017multimodal} from data which refer to data derived from different layers of the review writing process (from IP address at the lowest layer to review text at the highest layer). Twitter messages obtained from streaming API, certain user data (e.g., User ID, Tweet ID, and Time), Network Information (e.g., IP Address), and Device Information (e.g., GPS position). Chiu \textit{et al.}~\cite{chiu2017uncovering} characterized fraudsters as users with the same IP tweets who retweet with a different username for different purposes (political or commercial manipulation). 
% Datasets are tweets from twitter and DoS (Denial of Services) attacks. 
The multimodal data was then mapped to a 3D tensor, representing data evolution through time. The matrix in each time step represents the user ID, review ID, and the date of the written review as data modalities. 
% The contribution for this work is for Partially paired data which refers to data with incomplete entries. 
Chiu \textit{et al.}~\cite{chiu2017uncovering} provided an intuition of how modeling the behaviors of the user through time helps to improve fraudster detection.

To detect single fraudsters, Kumar \textit{et al.}~\cite{kumar2018identifying} proposed to first identify possible fraudster groups to find single fraudsters. This is based on the observation that behavioral clues on singleton reviews are scarce. To obtain information on a group of reviewers' collective behavior, a review-item matrix is constructed. The output of the proposed approach is a score given to each fraudster group. They achieved a recall of 88.79\% on the YelpZip dataset.

The graph Convolutional Network (GCN) was adopted by Wang \textit{et al.}~\cite{Wang2019FdGars} to detect fraudulent users in review platforms in a framework called FdGars. Wang \textit{et al.}~\cite{Wang2019FdGars} categorized fraudsters based on their motivations into three types: camouflaged users, crowdsourcing (group fraudsters), and fraudsters. FdGars performs both tasks of fraud and fraudster detection using review text statistics (e.g., review length and review symbol number) and user-based features (e.g., Review Quantity (RQ), Time-based Quantity Distribution (TQD), and Score-based Quantity Distribution (SQD)). 

\begin{table}
\centering
\caption{An overview of some of the features employed by Wang \textit{et al.}~\cite{Wang2019FdGars}. H/L depicts if a High/Low value of the feature is more likely to be associated with fraud.}
\label{tab:features-wang}
\begin{tabular}{m{2cm}m{3cm}m{0.5cm}m{1.5cm}}
\hline
Features &  Explanation & H/L & Formula\\
\hline
\hline
Review Length & Length (L) of review as the number of words & L & $L(R^i_u)$\\ \hline
Review Symbol Number & Number of non-alphabetic characters (NAC) in a review & H & $N_{NAC_u}$\\ \hline
Review Quantity & Number of reviews written by a fraudster & H  & 
$N_{R_u}$\\ \hline
Time-based Quantity Distribution & Distribution of the number of reviews over time & - & N/A\\\hline
Score-based Quantity Distribution & Distribution of the number of reviews over different ratings & - & N/A\\ \hline
\end{tabular}
\end{table}
Table~\ref{tab:features-wang} lists these features. The behavioral features connected users as nodes in a graph based on the users' proximity. The users' proximity was calculated based on the users' similarity in terms of extracted features. The GCN then labeled the reviews and users based on the similarity between them through a semi-supervised approach. FdGars achieved a performance of 93.8\% for F1-measure on a dataset collected from Tencent Inc.\footnote{\url{https://www.tencent.com/en-us}} with 302,097 reviews written by 82,542 users on 7,548 different applications. 
% Given the promising performance of GCN and multi-modal data, Adversary Situation Awareness (ASA) was proposed by Wen \textit{et al.}~\cite{Wen2020} as a GCN-based framework utilizing multimodal data. Wen \textit{et al.} claimed that the fraudsters employ camouflage strategy to escape detection. ASA used two different modalities: Activity, and Device. The former mainly focused on similar suspicious activities from fraudsters. The latter concentrates on the lower layers of the network such as the physical layer (MAC), IMSI (International Mobile Subscriber Identification Number), IMEI (International Mobile Equipment Identity), and transmission layer (IP). Such footprints disclose useful information e.g., users with the same IP and different user IDs are suspected to be fraudsters (or likely group fraudsters). Users with the same IMEI, IP, and MAC and different ids were connected to each other and the GCN was applied for the final labeling. The performance of the proposed approach by Wen \textit{et al.}~\cite{Wen2020} showed a recall of 90\% on a dataset collected from the Tencent Inc. website\footnote{\url{https://www.tencent.com/en-us}}. 
Recently, Danilchenko \textit{et al.}~\cite{danilchenko2022opinion} employed few shot learning in combination with a classical and a graph-based approach (specifically, belief propagation) and demonstrated that the combination of the two outperformed the individual approaches. Danilchenko \textit{et al.}~\cite{danilchenko2022opinion} first created a network of users connected to each other based on a set of products they have co-reviewed. To obtain the ground truth labels for reviewers, the proposed approach employed active learning via few shot learning. The potential of each user to be benign or fraudsters was set to [0.5,0.5]. Each user was represented with a feature vector based on features such as Positive Ratio (PR), Rate Deviation (RD), etc. The experiments showed that the proposed approach outperformed the previous study with an AUC of 72\% on the Yelp dataset.

In summary, the GCN provided a great opportunity for researchers to characterize the behavior of users in social review platform. Recently, multimodal data was also used to aggregate different aspects of user activity and hence improved the performance of fraudster detection.
% Fig. \ref{fig:single-user} shows a timeline of studies on single fraudster detection.  

\subsubsection{User Groups}
\label{sec:user-group}
% As explained in Sec.~\ref{sec:defs}, 
% the undeniable effectiveness of fraud reviews in social review platform encouraged the fraudsters to employ advanced strategies to attack businesses. 
% one strategy to maximize the impact of fraud is to get fraudsters work in groups. 
A fraudster group refers to a group of users coordinating to attack the same item. Most studies focused on employing graphs to model a group of users. Candidates of a potential fraudster group are first generated followed by a refinement step to improve the group's constitution.\\
% Graph-based models are largely used for fraudster group detection. 
Graphs in the fraud detection task are typically categorized into three sub-categories based on the node type:
\begin{itemize}
    \item \textbf{Monopartite:} The nodes are from the same type (users, or reviews)~\cite{shehnepoor20,Shehnepoor2017,Zhu2019}
    \item \textbf{Bipartite:} The nodes are from two different types (both users and reviews, or users and items)~\cite{Akoglu2013}
    \item \textbf{Multipartite:}  The nodes are from more than two different types (users, reviews, and items)~\cite{Shebuit2015}.
\end{itemize}
The connections between the nodes are defined based on the downstream task.

\textit{NEST} and \textit{BEST} were proposed by Zhu \textit{et al.}~\cite{Zhu2019} to address the overlooked small dense groups (modeled as a complete graph in the fraudster group detection task). A bipartite graph was utilized to model the users as a node type and users' activities as another node type. Users' activities are represented based on their social profile. Users with similar activities are more probable to be in a small group of fraudsters, representing explicit relations. Implicit relation was then defined as a relation between two users obtained through other users with similar activities. NEST and BEST both employ an iterative scorer to score the groups. BEST achieved an F-measure of 71\% on the Yelp dataset.\\
% With the success of graph-based models in modeling fraudster groups, a graph-based method was proposed by Zhao \textit{et al.}~\cite{Zhao2019} to address the limitations of previous studies in considering the global network features missed in a group representation. Label Propagation Algorithm (LPA) was incorporated to propagate the labels for the unlabeled users. 
% Each user as a node was connected to other users with a specific weight, representing the nodes' proximity. The nodes' connection is defined based on the phone call between 
% users. The weight of a relation between two users was determined in sequential iterations and the label of a user was determined based on the weights and labels of adjacent neighbors. The proposed approach by Zhao \textit{et al.}~\cite{Zhao2019} achieved an AUC of 87\% on the Lending dataset.

Bitarafan \textit{et al.}~\cite{Bitarfan2019}  employed a Heterogeneous Information Network (HIN) to utilize the merits of a graph in both the feature and classification step. The proposed approach; $SPGD\_HIN$; employed a bi-connected candidate group detection to detect the users with at least one co-reviewed item.

\begin{table}
\centering
\caption{An overview of some of the features employed by~\cite{Bitarfan2019}. H/L depicts if a High/Low value of the feature is more likely to be associated with fraud.}
\label{tab:features-bit}
\begin{tabular}{m{2cm}m{3cm}m{0.5cm}m{1.5cm}}
\hline
Features &  Explanation & H/L & Formula\\
\hline
\hline
Group Time Window & The time (T) window in which the members of a  group write the reviews & L & $Max([Max(T_{R_u}) - Min(T_{R_u})]\forall u\in group)$\\ \hline
Group Product Tightness & Number of common products with reviewers in a group& H & $N_{R_{u-p}\bigcap R_{u'-p}}$\\ \hline
\end{tabular}
\end{table}

The weight between two users was therefore computed based on the time-interval between the reviews written by the users. The users with strong ties (similar rating, and reviews in a time burst) were defined as a  bi-connected fraudster groups. Several features were extracted such as Group Time Window (GTW) and Group Product Tightness (GPT). Table~\ref{tab:features-bit} describes these features. Next, the metapath concept was employed to obtain the final weight between each of the two groups. Groups were labeled using a semi-supervised approach based on the calculated weights. The results yield a precision of 70\% on the Yelp dataset.
Similar to single fraudster detection multimodal data were incorporated alongside graph-based methods to improve the detection task.\\ 
\textit{GraphRfi}, proposed by Zhang \textit{et al.}~\cite{Zhang2020gcn}, utilized different features such as the number of rated products and the length of a username. Extracted features were combined with the rating of each review into a single vector and then fed to a Multi-Layer Perceptron (MLP) for the final prediction. \textit{GraphRfi} showed an F-measure of 99\% on the Yelp dataset. 

In summary, the earlier studies on fraudster group detection mostly relied on graph-based techniques to model the relations between members in a group, while later researches employed graphs to learn the vector representation of users. Most recent studies focused on multimodal data to better represent a group and for an improved performance.

\subsubsection{Bot Fraudsters}
\label{sec:generation}
% With advancements in deep learning, fraudsters have recently employed  bots to generate fraud reviews. 
As deep learning has advanced, fraudsters have recently started using bots to generate fraudulent reviews. In contrast to human-written fraud reviews, bot-generated reviews can be produced in a high volume, cutting the cost of human fraudster employment. Most importantly, bot-generated reviews leave no behavioral trace, as such bots can manipulate the behavioral clues (e.g., number of written reviews in a time burst, date of written reviews, the volume of reviews, review rating, etc.), which are often used in behavioral feature extraction. In this section, we explore the studies on the newly introduced topic of bot-generated review detection. 
% Written reviews can  be produced in a limited number and each review is paid by the businesses. In addition, Progress in deep learning provides a great opportunity for fraudsters to use DNN as a tool to generate fraud reviews indistinguishable from written ones. 
% Such generated reviews are easier to be generated in high volume with no cost. Finally, The generated reviews leaves no trace to be tracked down as a human behavior indicator. Consequently, generating such reviews comes with more benefits for businesses. In this section, we consider the bots as a user who generate fake reviews. With no previous studies on bot detection, the main focus here is on finding reviews generated by bots.
Bot-generated attacks were first investigated by Yuanshun \textit{et al.}~\cite{Yuanshun2017} using a pattern analysis method on the generated texts. Yuanshun \textit{et al.}~\cite{Yuanshun2017} proposed an approach to generate reviews based on the character distribution modeling with a simple Recurrent Neural Network (RNN). The generation process followed three consecutive steps. In the \textbf{first step}, the RNN was trained using characters instead of words, due to the low computational cost and the memory required for the training. \textbf{Next}, reviews were generated by the model. 
The limited size of the model led to information loss during back propagation. Hence, in the \textbf{third step}, the generated reviews were customized by replacing certain words with domain terminology. 

\begin{figure*}
    \centering
    \includegraphics[width= \linewidth]{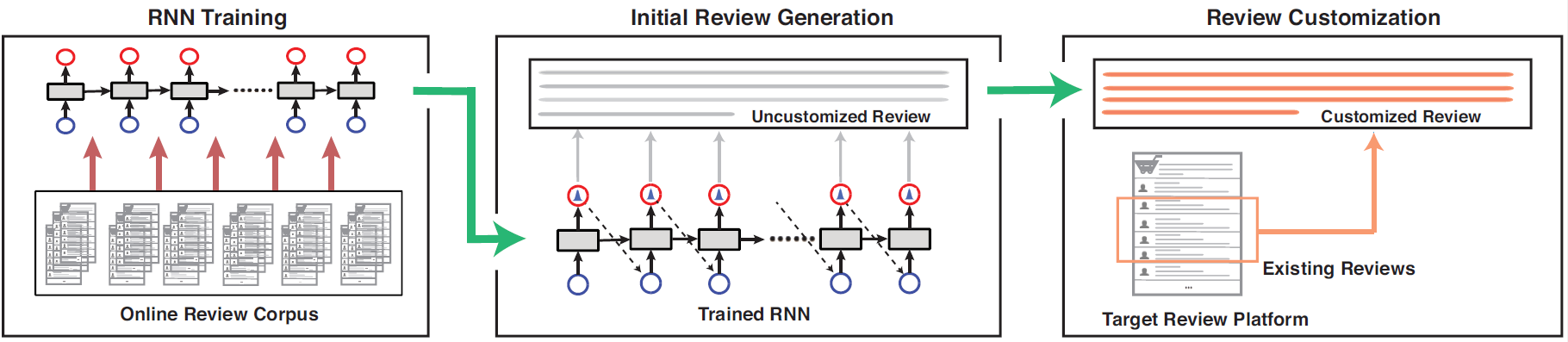}
    \caption{The overall framework proposed by~\cite{Yuanshun2017}.}
    \label{fig:yuan-framework}
\end{figure*}

% Fig.~\ref{fig:yuan-custom} shows an example of review customization by~\citet{Yuanshun2017}.
% \begin{figure}
%     \centering
%     \includegraphics[width=\linewidth]{samples/Figs/Yuan-custom.png}
%     \caption{Example of review customization by \citet{Yuanshun2017}.}
%     \label{fig:yuan-custom}
% \end{figure}
To detect generated reviews, an RNN classifier was trained based on the character distribution of the review test. To this end, two separate RNNs were trained, one with the generated reviews, and the another with the real reviews. The reviews in the test set were then fed to both models and the probability of each review being a fraud or real was calculated through a log-likelihood measurement. The likelihood was using a similar approach to that of the Siamese network where the outputs of two different trained neural networks are given to a scorer to calculate the final score. 
Fig.~\ref{fig:yuan-framework} shows the framework proposed by~\cite{Yuanshun2017}. 
The results of the proposed approach showed an F-measure of 86\% on the Yelp dataset.

Afterwards, several studies utilized Generative Adversarial Networks (GAN) to simulate the process of the review generation through bots. 
GAN was proposed by Goodfellow \textit{et al.}~\cite{Goodfellow2014a} to originally generate captions for images. The GAN consists of two main blocks: a Generator and a Discriminator. In bot review generation/detection context, before training the blocks, the generator and the discriminator are pre-trained using the samples from the training set derived from the TripAdvisor dataset. In the adversarial training phase, the pre-trained generator produces negative samples for the discriminator. 
% The generated samples are fed to the discriminator as the negative samples. 
In a zero-sum game between the two blocks, the generator generates fake samples and the discriminator learns to discriminate between real samples and fake ones. Both models are updated based on the backpropagation loss from the discriminator. Such a structure makes the GAN a promising choice for both bot generation and fraud detection tasks. 

Aghakhani \textit{et al.}~\cite{Aghakhani2018} proposed FakeGAN which employed a generator and two Convolutional Neural Networks (CNNs) as discriminators, one to discriminate between the generated reviews and human written fraud reviews, and the other one to discriminate between the set of fraud reviews (the combination of generated reviews and human written fraud reviews from the dataset) and real reviews. Two discriminators were proposed to address the limitation of the GANs in handling the ``mode collapse" problem. The mode collapse problem refers to a situation where the generator generates the reviews only from a single mode of distribution due to the multi-dimensionality of the data.

Before the adversarial training, the generator was pre-trained using Maximum Likelihood Estimation (MLE) and the discriminators were pre-trained using a cross-entropy loss function. FakeGAN was evaluated on TripAdvisor with 1600 reviews, containing 800 fraud reviews and 800 real reviews, and showed an accuracy of 78\% (See Sec.~\ref{sec:data}). 

To better simulate bot reviews, Shehnepoor \textit{et al.}~\cite{shehnepoor20} proposed ScoreGAN, a GAN-based approach where the generator utilized review ratings to improve the quality of the generated reviews. 
% ScoreGAN employed Information Gain Maximization (IGM) to generate reviews with certain ratings.
Their objective function introduced an additional term to maximize the correlation between a generated review for a given rating. 
% Hence, the generator can generate reviews with specific rating. 
Such a generator improves the quality of the generated reviews, which ultimately enhances the detection accuracy. 
% The detection of bot-generated reviews was significantly improved. 
% Furthermore, the proposed approach was also shown to be a significant improvement in terms of stability and scalability on two different datasets. 
ScoreGAN showed an accuracy of 84\% and 77\% on the Yelp and TripAdvisor datasets, respectively.
In summary, although only a few studies investigated bot review identification as fraud detection, recent approaches have demonstrated promising results. The use of GAN significantly improved the detection performance through bot review generation and the simultaneous training of one or two discriminator(s) to detect bot-generated reviews. 
% Fig.~\ref{fig:bot-user} is a timeline of the previous studies on the bot review detection.

\subsection{Item}
% Fraudsters using social review platform generally target a specific business and write reviews to promote/defame the items of the business. 
Some services or products may be more prone to fraud review attacks than others, referred as targeted items. As soon as a new product is introduced into markets, fraudsters may start to propagate fraud reviews. Detecting targeted items, therefore, is as important as fraudster and fraud review detection. Target item detection can indirectly help improve fraudster and fraud review detection. However, few studies investigated target item detection.
% studies took the business side to look at fraud detection phenomenon. 

% As one of the first studies on targeted item detection, \citet{wang2011review} incorporated the review rating in a Bayesian network to determine the ``trustiness" of a user, ``reliability" of an item, and ``honesty" of a review. The trustiness of a user is dependent of summation of the honesty scores of all written reviews by the user. The honesty of a review and the reliability of an item is calculated based on a proposed formulation. The scores are updated in a proposed Bayesian iterative algorithm. To evaluate the performance of the proposed approach three human assessors were employed, and a \textit{kappa} measure was used to output the final judgment of their evaluation. The evaluation showed a \textit{kappa}~\cite{fleiss1973equivalence} (an inter-evaluator agreement
% measure for any number of evaluators) measure of 60.3\% among the human judges on a dataset from the Resellrating\footnote{\url{www.resellerratings.com}} store.
% In latter these concepts are represented as trustiness of user, reliability of product, and honesty of review which are almost the same. 
% \citet{Mukherjee2013} employed a logistic function to model the components interacting with each others in a store dataset. 
Given the significance of item classification in fraud detection, 
% As discussed in Sec. \ref{sec:user-group}, fraudster groups target a certain business to collectively defame the product and decrease the product fame among users. 
FraudEagle, proposed by Akoglu \textit{et al.}~\cite{Akoglu2013}, uses 
% FraudEagle maps the users and items to a bipartite graph (as nodes) with signed edges representing the rating by a user to an item. The graph is depicted in Fig.~\ref{fig:fraudeagle-graph}.
% \begin{figure}
%     \centering
%     \includegraphics[width=\linewidth]{samples/Figs/fraudeagle-graph.jpg}
%     \caption{Example of review customization by \citet{Yuanshun2017}.}
%     \label{fig:fraudeagle-graph}
% \end{figure}
Loopy Belief Propagation (LBP) to propagate the fraudulent score through users and items. LBP is an iterative message passing, shown to be effective for various real-world applications~\cite{Yedidia2003}. As existing datasets contained no groundtruth, Akoglu \textit{et al.}~\cite{Akoglu2013} removed the users with a high probability of being fraudsters and then measured the items' rating difference before and after removing the suspected groups.
Akoglu \textit{et al.}~\cite{Akoglu2013} concluded that fraudsters significantly affected the average rating given to target items. Thus, the target items were identified by removing fraudsters while observing the impact on the SoftWare Marketplace dataset (SWM)\footnote{\url{http://odds.cs.stonybrook.edu/swmreview-dataset/}}.

Another work that uses item features is SPeagle~\cite{Akoglu2013}. To improve the fraudster detection performance, SPeagle~\cite{Shebuit2015} used both metadata and text to extract features from reviews, users, and items. 
The extracted vector (the combination of behavioral features such as negative ratio, positive ratio, and linguistic features such as content similarity) from review, user, and item was used to calculate prior knowledge for each component. The prior knowledge was then utilized to initialize a representation of each component in a multi-partite graph. 
Nodes in the graph represent user and items. Each edge represents a review written by a user on an item. LBP was employed to predict fraud reviews, fraudsters, and target items. SPeagle showed an AUC of 69.07\% and an AP of 42.45\%, on the Yelp dataset.

Ji \textit{et al.}~\cite{ji2020burst} utilized target item detection to improve fraudster group detection performance. Ji \textit{et al.}~\cite{ji2020burst} employed the item-related features (product rating  distribution, product average rating distribution, and suspicious score) to detect candidate fraudster groups in the Amazon dataset\footnote{\url{https://jmcauley.ucsd.edu/data/amazon/}}. To this end, the fraudster group features (group rating deviation, group size, group review  tightness, group one-day reviews, group extreme rating ratio, group co-Activeness, and group co-active review ratio), individual fraudster features (ratio of extreme rating, rating deviation, the most reviews one-day, review time interval, account duration, and active time interval reviews), and item related features were first extracted. Then, a threshold was applied to the overall score, resulting in a representation to initially determine the targeted items. A Kernel Density Estimation (KDE) method was then applied to calculate the burstiness of items. 
KDE is a technique that could asymptotically converge to any density function with sufficient samples. As such, KDE was employed to model the review sequence of a target item to find groups of reviewers in a review burst. A review burst refers to a sudden increase in the popularity of a product due to a potential fraud attack. Next, candidate groups were determined from the KDE output.
The items with burstiness value over a threshold were considered target items. 
% The candidate groups were then purified from genuine users. 
The approach demonstrated an F1-measure of approximately 75\% on the Amazon dataset. 

\subsection{Summary}
\begin{figure*}
    \centering
    \begin{tikzpicture}[scale = 0.5,timespan={}]
    \timeline[custom interval=true]{2017,...,2021}     
    \begin{phases}
        \initialphase{involvement degree=2cm,phase color=red!40!white}
        \phase{between week=1 and 2 in 0.5,
          involvement degree=1.5cm,phase color=red!70!white}
        \phase{between week=2 and 3 in 0.5,
          involvement degree=1cm,phase color=red!100!white}
        \phase{between week=3 and 4 in 0.5,
          involvement degree=3cm,phase color=red!70!black}
        \phase{between week=4 and 5 in 0.5,
          involvement degree=3cm,phase color=red!40!black}
      \end{phases}
        % \initialphase{involvement degree=2cm,phase color=blue}
        % \phase{between week=1 and 2 in 0.5,
        %   involvement degree=1.5cm,phase color=green!50!black}
        % \phase{between week=2 and 3 in 0.5,
        %   involvement degree=1cm,phase color=red!40!black}
        % \phase{between week=3 and 4 in 0.5,
        %   involvement degree=3cm,phase color=red!90!black}
        % \phase{between week=4 and 5 in 0.5,
        %   involvement degree=3cm,phase color=red!40!yellow}
    
    % 2016
      \addmilestone{at=phase-0.75,direction=75:2.2cm,
        text={Jindal \textit{et al.}}, text
        options={above}}
      \addmilestone{at=phase-0.135,direction=135:1.2cm,
        text={Wang \textit{et al.}}, text
        options={above}}
      \addmilestone{at=phase-0.225,direction=225:2.2cm,
        text={Akoglu \textit{et al.}}, text 
        options={below}}
      \addmilestone{at=phase-0.270,direction=270:3cm,
         text={Shebuit \textit{et al.}}, text options={below}}
    % 2017
      \addmilestone{at=phase-1.270,direction=270:1.7cm,
        text={Yuanshun \textit{et al.}}, text 
        options={below}}
      \addmilestone{at=phase-1.135,direction=135:0.7cm,
        text={Chiu \textit{et al.}}, text 
        options={above}}

    % 2018
      \addmilestone{at=phase-2.135,direction=135:2.2cm,
        text={Aghakhani \textit{et al.}}, text
        options={above}
        }
        
    % 2019
      \addmilestone{at=phase-3.90,direction=90:1.7cm,
        text={kauffmann \textit{et al.}}, text
        options={above}
        }
      \addmilestone{at=phase-3.135,direction=135:1.7cm,
        text={Barbado \textit{et al.}}, text
        options={above}
        }
      \addmilestone{at=phase-3.45,direction=45:1.7cm,
        text={Wang \textit{et al.}}, text
        options={above}
        }
      \addmilestone{at=phase-3.200,direction=200:1.7cm,
        text={Zhu \textit{et al.}}, text
        options={below}
        }
      \addmilestone{at=phase-3.230,direction=230:2.2cm,
        text={Zhao \textit{et al.}}, text
        options={below}
        }
      \addmilestone{at=phase-3.270,direction=270:2.2cm,
        text={Bitarafan \textit{et al.}}, text
        options={below}
        }
        
    % 2020
      \addmilestone{at=phase-4.15,direction=15:3.7cm,
        text={Shehnepoor \textit{et al.}}, text
        options={above}
        }
      \addmilestone{at=phase-4.60,direction=60:2.2cm,
        text={Zhang \textit{et al.}}, text
        options={above}
        }
      \addmilestone{at=phase-4.225,direction=225:0.7cm,
        text={Ji \textit{et al.}}, text
        options={below}
        }
      \addmilestone{at=phase-4.270,direction=270:1.2cm,
        text={He \textit{et al.}}, text
        options={below}
        }
      \addmilestone{at=phase-4.315,direction=315:4.2cm,
        text={Ruan \textit{et al.}}, text
        options={below}
        }
      \addmilestone{at=phase-4.345,direction=345:2.7cm,
        text={Wen \textit{et al.}}, text
        options={below}
        }

    \end{tikzpicture}
    \caption{The volume timeline of the studies on different components in fraud review. Diameter of each circle is proportional to the number of studies for one year period [2017-2018, 2018-2019, etc.]}
    \label{fig:componentsTime}
\end{figure*}
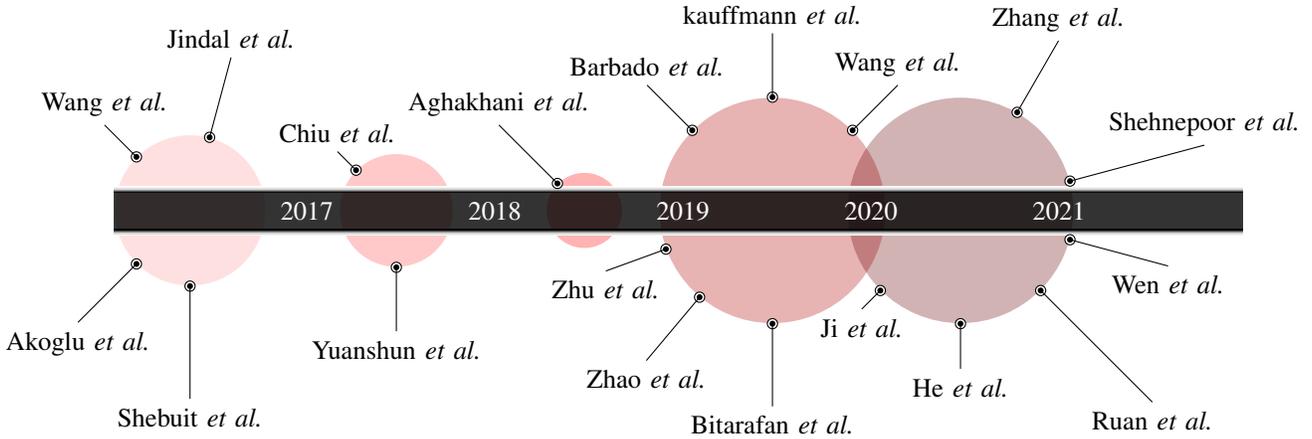

To better illustrate the evolution of research on component detection, we have provided a timeline of previous studies on various components in Fig.~\ref{fig:componentsTime}. 
% With the advancement of new techniques, more researchers are concentrating on components as the main topic to consider. 
As new techniques have advanced, more researchers have focused on components as the primary subject of study. Hence, Fig.~\ref{fig:componentsTime} shows a boost toward studies focusing on each of the three components (especially between 2019-2021).  

\section{Second Step: Feature Extraction}
\label{sec:features}
\begin{figure}
    \centering
    \includegraphics[width=\linewidth]{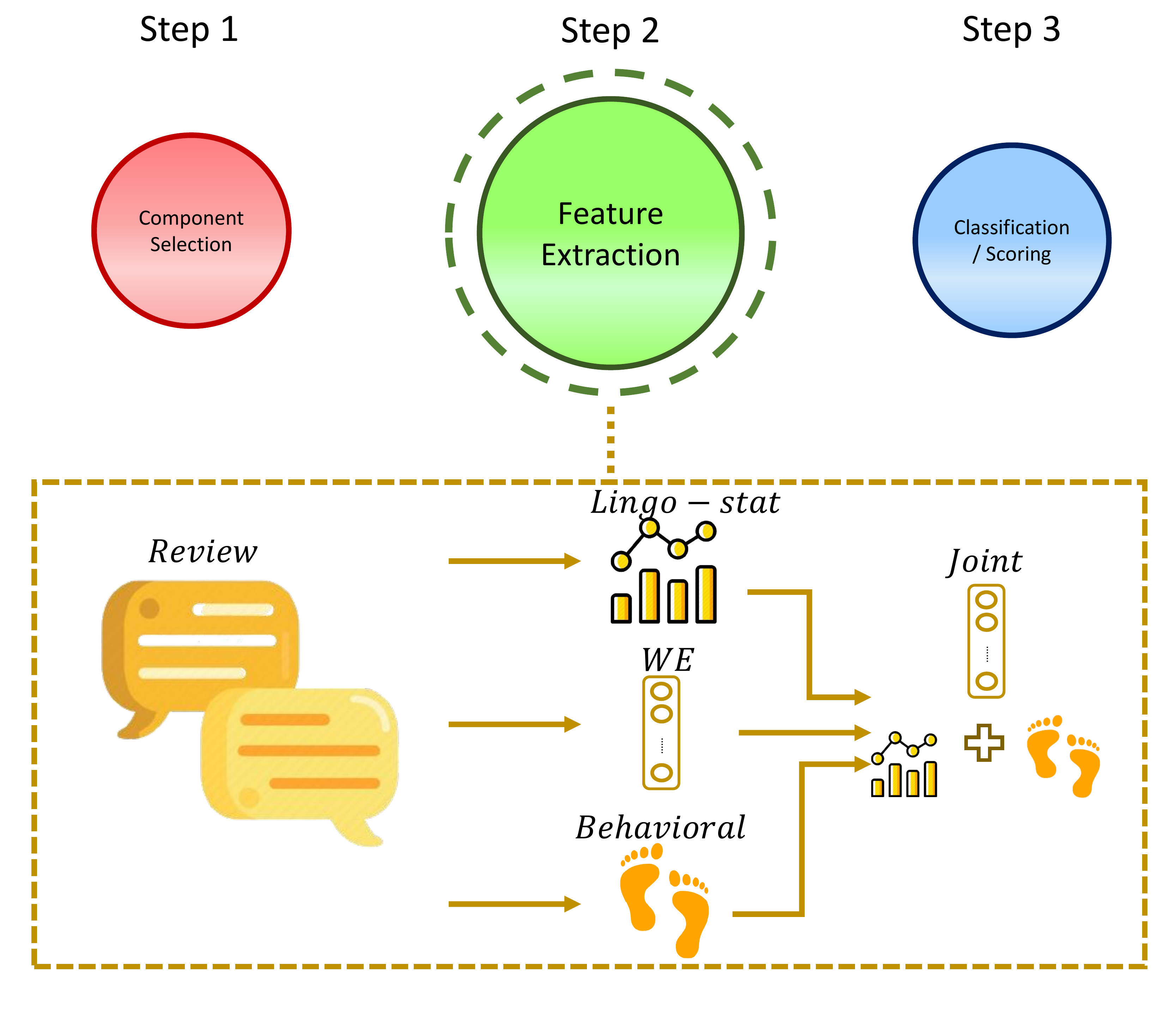}
    \caption{An overview of the feature extraction stage in a fraud detection framework.}
    \label{fig:features-step}
\end{figure}

After choosing a component of study, features of the selected component need to be extracted,
% For users and items, data are aggregated from all available reviews.% For an item, all the reviews written for that item are collected, and for a user all the reviews written by that user are combined. Obviously, we require only one review to form the data for a specific review, as a component. 
either from the review text or metadata. Features extracted from the review text are referred to as text-based features, while features making use of metadata are referred to as behavioral features. 
% Early studies utilized the lingo-statistic text-based features, while more recent studies employed Deep Learning to extract features as ``Word Embeddings". 
To achieve a better representation, text-based features and behavioral features can be concatenated or jointly learned through deep learning to obtain \textit{joint} features. Fig.~\ref{fig:features-step} displays the detailed steps involved in the feature extraction stage.

\subsection{Text-based} 
Features extracted from the text for fraud detection can be grouped into two categories, namely lingo-statistics features and vector space features (i.e., word embeddings). 
% Early methods extracted lingo-statistic text features and used them either separately or in combination with behavioral features. Word Embeddings of review text were recently employed to extract a more effective representation of words that are called word embeddings (WEs). ``Word2Vec" is one of the first embedding methods, with the more recent ones such as BERT~\cite{BERT} concentrating on context-based embeddings.
%through the ``word2vec" methods. In the following, we represent and discuss these two categories. 

\subsubsection{Lingo-Statistics}
\label{sec:statistical}
% lingo-statistic features refer to the features extracted to describe the . Statistical text-based features are known to be traditional features, and were mostly used in early studies on fraud detection. \\
Lingo-statistics text-based features were proposed by Lai \textit{et al.}~\cite{Lai2011} to detect different types of \textit{Spam} contents, including fraud reviews. As one of the first text-based features, a unigram language model was developed based on Part of Speech (POS) tagging. To calculate the similarity between two different reviews, the Kullback-Leibler (KL) divergence measure was used, where the likelihood of a review generating the contents of the other review was used. The performance of the proposed approach was evaluated on the TREC dataset and showed a precision of 94\%.

Mukherjee \textit{et al.}~\cite{mukherjee2013fake} also employed Part Of Speech (POS) tagging to extract the bigram and unigram features from review text based on a specified frequency. An SVM model was trained on the extracted POS features and showed an accuracy of 85.1\% on the Yelp dataset.   
\\

Banerjee \textit{et al.}~\cite{Banerjee2015} extended the concept of the lingo-statistic text-based features using \textit{understandability}, \textit{level of details}, \textit{writing style}, and \textit{cognition indicators}. For each, a set of features was proposed to describe the review. Several classifiers were employed such as Logistic Regression (LR), Decision Tree (DT), Neural Network (NN), Naive Bayes (NB), Random Forest (RF), Support Vector Machine (SVM), and voting. A dataset of 900 genuine reviews was collected from different hotel websites including \textit{Hotels.com}, \textit{Expedia.com}, and \textit{Agoda.com}. To provide positive samples, 60 fake reviews were also deliberately written by experienced reviewers. LR yielded the best performance with an AUC of 81.5\% on the Expedia dataset.
Though singletone text-based features (i.e., features from a single review) provide a good level of characterization, the similarity between reviews (known as pair-wise features) provided a more powerful means for fraud detection and is considered as one of the most important lingo-statistic text-based features. 

% Xu \textit{et al.}~\cite{xu2015combating} categorized the similarity-based features as pairwise features (such as product-based time deviation, product-based text similarity, and brand-based sentiment deviation). 

% To predict the review class, a Mutual Disclosure Model is employed. Mutual Disclosure Model is a label propagation model, which obtains the final label through sequential iterations. The idea of Mutual Disclosure Model is very similar to LBP~\cite{Akoglu2013,Shebuit2015}, where each node is first assigned with a word embedding based on the features. Such vectors are then used as a prior knowledge in a label propagation algorithm to refine the probability of each user, involved in a fraudster group activity or not. The results showed a Normalized Discounted Cumulative Gain (NDCG) of 72\% on the Amazon dataset. 

With many lingo-statistic text-based features proposed in the early years,  Crawford \textit{et al.}~\cite{crawford2016reducing} decided to evaluate the importance of each word in a review for the fraud detection task. To this end, Crawford \textit{et al.}~\cite{crawford2016reducing} designed an approach to determine the keywords for the fraud detection task. Several feature selection techniques were utilized to determine the importance of each word in a review text for the fraud detection task. The techniques included word frequency, Chi-Squared (CS), and Mutual Information (MI). Then the presence of the 100 most frequent words in a review was given as a feature vector. The word vector was then fed to different classifiers such as LR, NB, SVM, etc. The best result was achieved for SVM on the Mechanical Turk (MT) dataset with an AUC of 87\%. 

As the most recent study employing lingo-statistic features, Abri \textit{et al.}~\cite{abri2020linguistic} extracted new linguistics features such as lexical diversity, emotiveness, etc., to detect fraud reviews. To evaluate the performance of different classifiers, the extracted features were fed to an SVM, Naive Bayesian, Random Forrest, etc. The best performance was demonstrated by the Multi-Layer Perceptron (MLP) with an accuracy of 79.09\% on reviews crawled from online restaurants. 
The obvious shortcoming of lingo-statistic text-based features is that indicators researchers adopted can easily be manipulated by fraudsters. However, some lingo-statistic text-based features such as language models, are employed to reduce the need for explicit indicators to deal with fraud detection. A language model can be employed in different ways as lingo-statistic features. Lai \textit{et al.}~\cite{Lai2011} employed the unigram model to calculate the likelihood of each review is a fraud, while Shebiut \textit{et al.}~\cite{Shebuit2015} employed the language model to represent each review as a bag-of-bigrams and then calculate the similarity between two reviews. 
% Therefore, depending on the task, the language model is utilized to represent the review. 
With the advent of Deep Learning, different neural networks such as RNN were used to consider the language model in a review and improve the detection task. 
% Fig. \ref{fig:statistical-text} shows a timeline of previous studies on lingo-statistic text-based features.

\subsubsection{Word Embeddings}
\label{sec:VR}
Lingo-statistic text-based features suffer not only from sparsity and subjectivity but also from easy manipulation by fraudsters. On the other hand, the newly introduced Word Embedding (WE) techniques utilized vector representation of words to overcome the such limitation.
Yafeng \textit{et al.}~\cite{Yafeng2016} claimed that a genuine review contained more contextual embedding, i.e., it is more informative than fraud reviews, as demonstrated in previous studies~\cite{Xu2015}.

Accordingly, Li \textit{et al.}~\cite{Li2016} extracted different text-based features such as review content similarity (pair-wise features) between two different reviews and then extracted a WE for each one. Emotion modeling was proposed based on the sentiment of the review text. The sentiment diversity, i.e., means square deviation of the negative, positive, and neutral word ratios is obtained based on their squared difference. The combination of all text-based features was then fed to different classical classifiers for labeling. The performance showed an F1-score of 93\% on the Yelp dataset with a Decision Tree (DT) as the classifier, shown to be more effective than lingo-statistic features. 
Given the success of WE in extracting sentiment, researchers began to employ WE in the fraud detection task.
% Surrounding information plays an important role in deciding whether a review is deceptive or not, while in old works it has been claimed that word occurrence plays a more important role in determining whether an opinion is deceptive opinion or not.
Zhang \textit{et al.}~\cite{Zhang2018} claimed that the reviews written by fraudsters were different from real ones since fraudsters typically could not reflect their experience of using the product in their reviews, potentially due to the lack of experience with the products. 
% As such, mapping reviews to Word Embed- dings would improve the classification task. 
% Zhang \textit{et al.}~\cite{Zhang2018} employed a Recurrent Convolutional Neural Network (RCNN) to map each word into a 6-component feature space. The skip-gram WE~\cite{Mikolov-skip} is utilized alongside six extracted components. The first two components were derived from training the RCNN with fraud and genuine reviews. The third and fourth components were obtained by training the RCNN with the left words surrounding each word in deceptive and real reviews. The fifth and sixth components were also derived from training the RCNN with the right words surrounding the target word. 
% This work also claims if the length of word vectors be less than 50, then since the word is mapped into smaller feature space, then it cannot represent the word, properly and as a result the performance will be dropped, eventually. 
% The framework achieved an average accuracy of 85\% on the TripAdvisor dataset.  \\
% With deep learned end-to-end techniques introduced for different machine learning tasks, researchers utilized them for fraud detection. 

% In other words, with deep learning a review is simply fed to the detection algorithm as an input and the fraudulent score is provided as the output (end-to-end).  
Ren \textit{et al.}~\cite{Yafeng2016} demonstrated that the Recurrent Convolutional Neural Network (RCNN) outperformed traditional classifiers.
%, with stronger hidden layers to extract latent aspects in a text. 
% and turn them into combined feature values which traditional feature fails to do so. 
The RCNN takes distributed word embeddings as an input to learn a continuous document representation for each review. The Continuous Bag of Words (CBOW) word embeddings were obtained using Word2Vec pre-trained on the dataset and then fed to an RCNN for extracting a document representation. 
% Hereby, the behavioral features were used in combination with WE to improve the representation of a review document. 
Finally, the representation is fed to a softmax layer for classification. The results showed an accuracy of 83.6\% on the Mechanical Turk dataset.
Jia \textit{et al.}~\cite{Jia2018} also employed term frequency and LDA (Latent Dirichlet Allocation) alongside WE. LDA is an approach that extracts review topics by highlighting important words (the most frequent ones). To calculate the term frequency, 5000 of the most frequent words (using unigram) were extracted from reviews and each word’s frequency was calculated. The Word Embeddings (WE) were pre-trained for each word using the skip-gram model~\cite{Mikolov-skip} and then an average of the words' embeddings was used as a vector representation of a review document. Jia \textit{et al.} exploited 5 topics for fraud and genuine reviews and each topic was described in 8 words. Such a topic modelling helped to find the most popular trends among fraudsters and genuine users. A Multi-Layer Perceptron (MLP) was applied to the final representation to achieve an accuracy of 81.3\% on the Yelp dataset.

In summary, the most recent studies employed word2vec techniques for fraud detection to extract WE for reviews. The proposed approaches extracted WE either independently from the classification phase in a pre-training phase (e.g., skip-gram, Glove, etc.) to obtain the representation, or employed an end-to-end deep learning technique (e.g., RCNN). To improve the performance of WE in improving fraud detection accuracy, the representations were also combined with behavioral features. As the most recent approach, new pre-trained architectures are used to provide finetuned representations, which can be easily transformed to a different scope. Bidirectional Encoder Representations from Transformers (BERT)~\cite{BERT} has been used to bi-directionally learn the representations using transformers. The key difference is that BERT considers both directions for learning the embeddings instead of either left or right. Such techniques can be helpful for obtaining a better representation in the fraud detection task. 
% A timeline of recent studies employing WE as feature is provided in Fig.~\ref{fig:vr-text}.

\subsection{Behavioral}
\label{sec:behavioral}
Behavioral footprints play an undeniable role in detecting fraud review. Such features can be used to represent a user, a review, or an item.

With the demonstrated importance of rating as a behavioral feature in early studies, Luca \textit{et al.}~\cite{Luca2016} employed the rating from reviews and applied a regression model to determine the fraudulent score of the reviews. 
% The observations for this study demonstrates increasing competition between companies leads to a considerable expansion of spam reviews on products. 
As the review rating was the only feature employed by Luca \textit{et al.}~\cite{Luca2016}, the regression ($R^2$) value between review ratings and predictions was calculated to evaluate the performance of the proposed approach. The results demonstrated a regression of 0.68 between the review rating and the probability of the review is a fraud on the Yelp dataset. 

Goswami \textit{et al.} \cite{goswami2017impact} proposed an approach with user-based features as the input to an artificial neural network (ANN) to spot fraud reviews on a dataset crawled from the Yelp website. Goswami \textit{et al.} employed different behavioral features such as photo count, compliment, friend count, tips, followers, and funny and useful votes as extracted features. To determine the importance of each new feature a decision tree was used. The framework showed a precision of 91.53\% and a recall of 99.98\% on a manually collected dataset.

To incorporate effective behavioral features, an Unauthorized Optimization based De-Anonymization (UODA) approach was proposed by Hernandez \textit{et al.}~\cite{Hernandez2018} to deal with fraudsters with multiple accounts on social review platforms.
% A method was employed to calculate the weights between two reviews on different items based on the discussed subjects in each review. 
\textit{Inter-review time} was also employed as a key feature in determining fraudsters (also referred to as \textit{burstiness} in different studies). \textit{Rating difference} between two reviews was used as another behavioral feature to reflect the correlation between two different reviews. An iterative classifier was then applied to the features extracted for each user to calculate the similarity between different users and then calculated the fraudulent score for each user. The UODA showed an F1-score of 93.84\% on the Google play dataset.

Shaalan \textit{et al.}~\cite{shaalandetecting} proposed a two-step approach concentrating on the key features of fraudsters such as adding redundant information in written review text or writing reviews in bursts.  In the first step,  a  Deep  Boltzmann Machine (DBM) was used as the aspect-level sentiment model. In the second step, a Long Short Term Memory (LSTM) was applied to the extracted sentiment aspect-level representation. The output of the LSTM was the label of a review to be fraudulent/genuine. The approach shows an accuracy of 80.85\% on the Yelp dataset.

In summary, despite being utilized in fraud detection since the early studies, behavioral features are still considered as one of the most effective indicators for the fraud detection task. However, behavioral features may have limitations in handling bot-generated reviews. Different modalities of data, such as IP, MAC, etc., may provide a better overview of users' activity in social review platforms. 
% Fig.~\ref{fig:behavioral} shows a timeline of previous studies employing behavioral features.
\subsection{Joint}
Joint features refer to the combination of text-based and behavioral features; through either a simple concatenation or a joint learning process. Joint features were proposed to improve the representation of the components in social review platforms.
% \citet{Li2011} proposed a joint representation of the text-based and behavioral features for all components in a social review platform (review, user, item). 

\begin{table}
\centering
\caption{An overview of some of the features employed by Kumar \textit{et al.}~\cite{Kumar2019}. H/L depicts if a High/Low value of the feature is more likely to be associated with the fraud.}
\label{tab:features-kumar}
\begin{tabular}{m{2cm}m{3cm}m{0.5cm}m{1.5cm}}
\hline
Features &  Explanation & H/L & Formula\\
\hline
\hline
Review Count & Number of reviews written by a user & H & $N_{R_u}$\\ \hline
Review Gap & The time gap between written reviews of a user & L & $T_{R^i_u}-T_{R^{i-1}_u}$\\ \hline
Entropy & The difference between users' reviews & L & N/A\\ \hline
\end{tabular}
\end{table}

% Several lingo-statistic text-based features were employed by \citet{Li2011} such as unigrams, bigrams, second-person pronouns ratio, number of objective and subjective words in a text, and the cosine similarity between two reviews. The text-based features were then concatenated with the Rating Deviation (RD) and fed to a naive Bayesian classifier to demonstrate an F1-score of 58.3\% on the Epinion dataset. \\
Given the promising performance of the joint representation, researchers recently employed the joint learning of the text-based and the behavioral representations to address challenging problems such as the cold start problem.

The cold-start problem was first investigated by Wang \textit{et al.}~\cite{wang2017handling} and refers to the limitation of the filtering algorithm to detect fraud reviews written by a new user. 
Wang \textit{et al.}~\cite{wang2017handling} was presumably the first study to address the cold-start problem. This is a challenging task since there is no information history of a new user, resulting in the failure of the detection algorithm to detect new fraudsters.

% This work is proposed to overcome the defects in previous works for this matter, including requirement for observing data in long term and extracting statistics from them. 
% In addition, filtering systems need to act at the moment to filter such contents and prevent them infect other reviewers. 
% Thus, designing algorithms that can detect fraud reviews ASAP is important. 
Wang \textit{et al.}~\cite{wang2017handling} claimed that features proposed by early studies fail to describe user behaviors. Therefore, a \textit{TransE} model was employed to encode a graph structure between an item, a user, and a review (as a head/translation/tail relation). Lingo-statistic text-based features (e.g., review length and maximum cosine similarity) and behavioral features (e.g., rating deviation) were extracted for components in the Yelp and Amazon platform. Continuous Bag of Words (CBoW) was employed as a Word Embedding technique to obtain the word embedding from the review text. The extracted features were then jointly learned through an objective function on both the text-based and behavioral features. Finally, a CNN was used to classify the users as genuine/fraudster on the Yelp dataset. 
The results demonstrated an accuracy of 58.3\% on the Yelp dataset.

To address the limitations of the framework proposed by Wang \textit{et al.}~\cite{wang2017handling}, an attribute-based framework, namely AEDA (Attribute Enhanced Domain Adaptive), was proposed. Three types of relations were defined: attribute-attribute, entity-attribute, and entity-entity. Pairwise features were extracted to better handle the cold-start problem and the objective function was defined based on the relations in the TransE model. 
% TransE is proposed in  \cite{bordes2013translating} where components in graph are represented as head, label/translation and tail $(h,l,t)$. For this purpose, tail to head representation plus a low dimensional representation related to label. So this transE model tries to model distance between $(h+l)$ and $(t)$ and learn $(t)$ from $(h+l)$ using some objective function. 
AEDA demonstrated an accuracy of 80.0\% on the Yelp dataset.  

% An extension to \cite{Kumar2018} is proposed by 
The joint representation learning of the text-based and behavioral features has also been broadly employed to address the data challenges in recent years. Kumar \textit{et al.}~\cite{Kumar2019} proposed an unsupervised approach to overcome the limitations of supervised learning. The proposed approach first fits the best distribution to the different behavioral features such as \textit{Review Count (RC)}, \textit{Review GAP (RG)}, and \textit{rating entropy} as shown in Table~\ref{tab:features-kumar}. The length of the review (i.e., the number of words) was also extracted as a lingo-statistic text-based feature.
% Table~\ref{tab:features-kumar} shows the features.
% The study first introduces the cases in which there exists a relationship between reviews. For example, the positive reviews on an item lead to more positive reviews.
To incorporate the inter-dependencies between reviews, extracted features were modeled as different Dirichlet distributions. 
The distributions were modeled using a Gaussian Mixture Model. According to Kumar \textit{et al.}, a fraudster was more likely to be deviant in a distribution compared to a genuine user. The proposed approach demonstrated an AUC of 70\% on the Yelp dataset.

As one of the most recent studies, Xiang \textit{et al.}~\cite{xiang2022deep} proposed a framework to represent linguistic information using BERT. Behavioral features such as the Maximum Number of Reviews (MNR), Positive Ratio (PR), Negative Ratio (NR), etc. were extracted from the reviews and categorized into two categories of product-based and user-based features. Such features were then fed to the Graph Convolutional Network (GCN). Both BERT and behavioral features were then fused into one global feature and fed to a Softmax layer for final classification. The proposed approach demonstrated an accuracy of 69.9\% on the Yelp dataset.

The joint representation extraction provided an opportunity to deal with more challenging problems recently introduced in social review platforms. 
However, the performance of a framework with a joint representation of components as the input highly depends on how the features are combined. 

\begin{center}
\begin{table*}[hbt!]
% \captionsetup{font=8p}
\centering
\caption{Studies based on feature categorization.}\label{tab:featurs}
\begin{tabular}{|p{1.5cm}|p{1.5cm}|p{4cm}|p{3cm}|p{2cm}|}
\hline
\multicolumn{2}{|c|}{Feature Categories}&  Features & Paper & Component \\
\hline
\hline
\multirow{8}{*}{Text-based} & \multirow{4}{*}{Lingo-stats} & 
Unigram language model & Lai \textit{et al.}~\cite{Lai2011} & Users\\ \cline{3-5}
& & 
POS-based features & Mukherjee \textit{et al.}~\cite{mukherjee2013fake} & Reviews \\ \cline{3-5}
& & 
Pairwise features & Xu \textit{et al.}~\cite{xu2015combating} & Users \\ \cline{3-5}
& & 
puasality, lexical diversity, emotiveness & Abri \textit{et al.}~\cite{abri2020linguistic}  & Reviews \\ \cline{2-5}
& \multirow{4}{*}{WE} & 
WE through skip-gram, positive and negative ratio & Li \textit{et al.}~\cite{Li2016}  & Reviews \\ \cline{3-5}
&  & 
CBOW & Ren \textit{et al.}~\cite{Yafeng2016}   & Reviews \\ \cline{3-5}
& & Skipgram & Zhang \textit{et al.}~\cite{Zhang2018}  & Users \\ \cline{3-5}
& & Skipgram + word frequency & Jia \textit{et al.}~\cite{Jia2018}  & Reviews \\ \hline
\multicolumn{2}{|c|}{\multirow{6}{*}{Behavioral}} & 
The Ratio of First Reviews, Rating Extremity, etc. & Mukherjee \textit{et al.}~\cite{Mukherjee2013} & Reviews\\ \cline{3-5}
\multicolumn{2}{|c|}{}& 
review rating& Luca \textit{et al.}~\cite{Luca2016} & Reviews\\ \cline{3-5}
\multicolumn{2}{|c|}{}& 
photo count, compliment, friend count, tips, followers, etc.& Goswami \textit{et al.}~\cite{goswami2017impact} & Reviews\\ \cline{3-5}
\multicolumn{2}{|c|}{}& 
inter-review time, rating difference, etc.
& Hernandez \textit{et al.}~\cite{Hernandez2018} & Users\\ \cline{3-5}
\multicolumn{2}{|c|}{}& 
review rating
& Li \textit{et al.}~\cite{Li2019} & Users\\ \cline{3-5}
\multicolumn{2}{|c|}{}& 
Burstiness
& Shaalan \textit{et al.}~\cite{shaalandetecting} & Reviews\\ \hline
\multicolumn{2}{|c|}{\multirow{5}{*}{Joint}} & 
unigrams, bigrams, second-person pronouns ratio, number of objective and subjective words in a text, and the cosine similarity between two review texts, rating deviation & Li \textit{et al.}~\cite{Li2011} & Reviews\\ \cline{3-5}
\multicolumn{2}{|c|}{}& 
inter-review time, rating difference, etc.
& Hernandez \textit{et al.}~\cite{Hernandez2018} & Users\\ \cline{3-5}
\multicolumn{2}{|c|}{}& 
Review Length (RL), Maximum Cosine Similarity (MCS)
& Wang \textit{et al.}~\cite{wang2017handling} & Reviews\\ \cline{3-5}
\multicolumn{2}{|c|}{}& 
rating difference, date difference, CBOW
& You \textit{et al.}~\cite{you2018attribute} & Users \\ \cline{3-5}
\multicolumn{2}{|c|}{}& 
Review Count (RC), Review GAP (RG), and Rating entropy, Review Length (RL)
& Kumar \textit{et al.}~\cite{Kumar2019} & Users \\ \cline{3-5}
\multicolumn{2}{|c|}{}& 
Maximum Number of Reviews, Positive Ratio, Negative Ratio, BERT & Xiang \textit{et al.}~\cite{xiang2022deep} & Reviews \\ \hline

\end{tabular}
\end{table*}
\end{center}

\subsection{Summary}
\begin{figure*}
    \centering
    \begin{tikzpicture}[scale = 0.4,timespan={}]
    \timeline[custom interval=true]{2017,...,2020}     
    \begin{phases}
        \initialphase{involvement degree=3cm,phase color=green!40!white}%9
        \phase{between week=1 and 2 in 0.7,
          involvement degree=1cm,phase color=green!80!white}%1
        \phase{between week=2 and 3 in 0.5,
          involvement degree=2cm,phase color=green!80!black}%4
        \phase{between week=3 and 4 in 0.5,
          involvement degree=1.5cm,phase color=green!40!black}%2
      \end{phases}
    
    % 2016
      \addmilestone{at=phase-0.165,direction=165:5.7cm,
        text={Lai \textit{et al.}}, text
        options={above}}
      \addmilestone{at=phase-0.150,direction=150:5.2cm,
        text={Banerjee \textit{et al.}}, text
        options={above}}
      \addmilestone{at=phase-0.140,direction=135:1.2cm,
        text={Xu \textit{et al.}}, text 
        options={above}}
      \addmilestone{at=phase-0.120,direction=120:3.2cm,
         text={Crawford \textit{et al.}}, text options={above}}
      \addmilestone{at=phase-0.185,direction=185:7.2cm,
         text={Mukherjee \textit{et al.}}, text options={below}}
      \addmilestone{at=phase-0.200,direction=200:3.2cm,
         text={Li \textit{et al.}}, text options={below}}
      \addmilestone{at=phase-0.230,direction=230:2.2cm,
         text={Li \textit{et al.}}, text options={below}}
      \addmilestone{at=phase-0.215,direction=215:5.2cm,
         text={Luca \textit{et al.}}, text options={below}}
      \addmilestone{at=phase-0.180,direction=180:6.2cm,
         text={Ren \textit{et al.}}, text options={above}}
    % % 2017
      \addmilestone{at=phase-1.120,direction=120:1.7cm,
        text={Wang \textit{et al.}}, text 
        options={above}}

    % % 2018
      \addmilestone{at=phase-2.90,direction=90:1.2cm,
        text={You \textit{et al.}}, text
        options={above}
        }
      \addmilestone{at=phase-2.30,direction=30:1.2cm,
        text={Jia \textit{et al.}}, text
        options={above}
        }
      \addmilestone{at=phase-2.225,direction=225:1.2cm,
        text={Hernandez \textit{et al.}}, text
        options={below}
        }
      \addmilestone{at=phase-2.315,direction=315:3.2cm,
        text={Zhang \textit{et al.}}, text
        options={below}
        }
        
    % % 2019
      \addmilestone{at=phase-3.45,direction=45:2.2cm,
        text={Li \textit{et al.}}, text
        options={above}
        }
      \addmilestone{at=phase-3.325,direction=325:1.7cm,
        text={Kumar \textit{et al.}}, text
        options={below}
        }

    \end{tikzpicture}
    \caption{The volume timeline of the studies on features employed in fraud review detection. Diameter of each circle is proportional to the number of studies for one year period [2017-2018, 2018-2019, etc.]}
    \label{fig:features-tl}
\end{figure*}
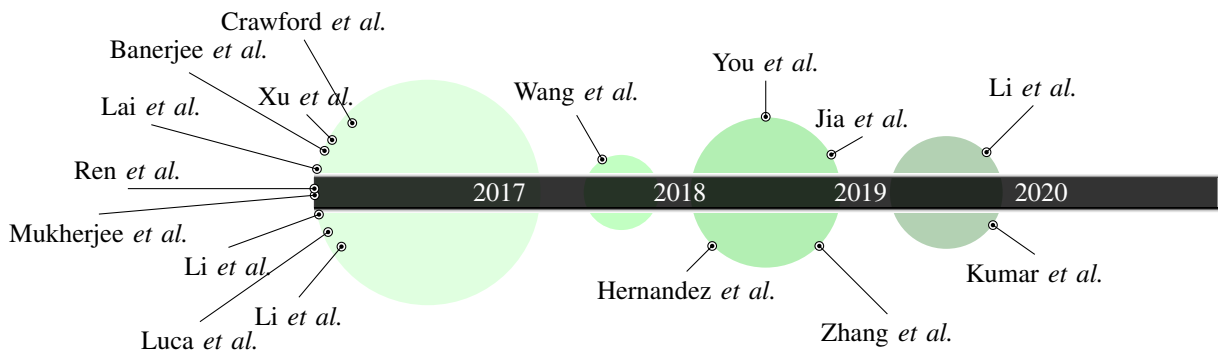

To provide a general overview of the evolution of features, a volume timeline is depicted in Fig. \ref{fig:features-tl}  showing the most recent studies on different types of features for fraud detection. Earlier studies (before 2017) focused on proposing new features, while deep learning provided the opportunity to use neural models to represent features in vector spaces.  Fig.~\ref{fig:features-tl} shows a boost in the number of studies that focused on features in 2018, with more studies concentrating on features in the subsequent years. A summary of employed features is shown in Table~\ref{tab:featurs}

\section{Third Step: Classification/Scoring}
\label{sec:approaches}
\begin{figure}
    \centering
    \includegraphics[width=\linewidth]{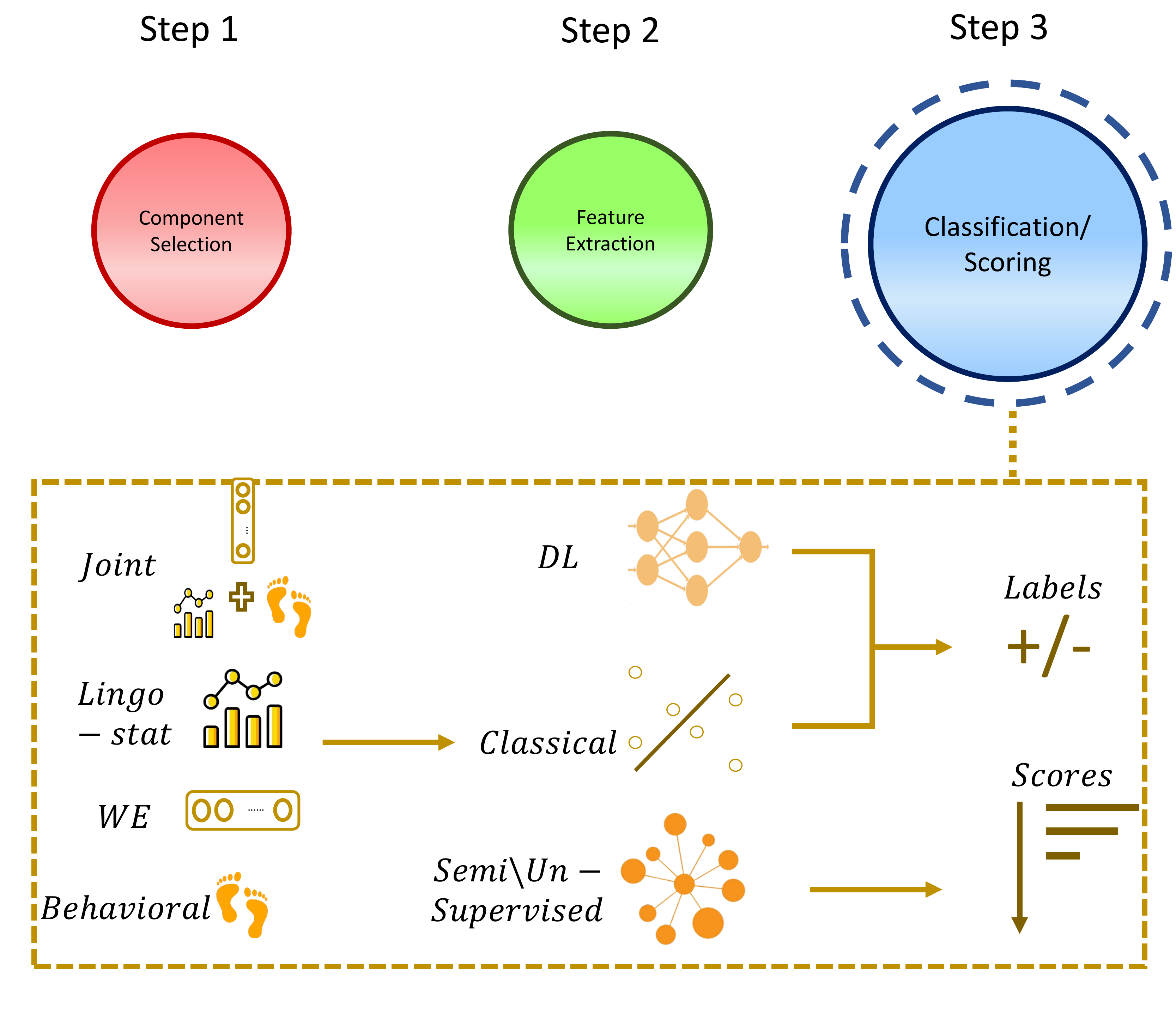}
    \caption{An overview of different modules, feature types, and outputs used in a fraud detection framework.}
    \label{fig:approach-step}
\end{figure}

In the final step of the proposed fraud detection framework, the extracted features or feature representations are used to classify each component. The approaches are categorized into three main sub-categories; supervised, semi-supervised and unsupervised. These categories can output the binary labels indicating the class of the component, or ranked scores showing the probability of the component belonging to a specific class. An overview of approaches is given in Fig.~\ref{fig:approach-step}.
% The output for the supervised approaches (including both deep learning and classical approaches) is a binary label.

% representing the class of each component. 
% On the other hand, the semi/un-supervised approaches output the probability of each component belonging to a specific class. The approaches are depicted in Fig.~\ref{fig:approach-step}.
\subsection{Supervised}
\label{sec:supervised}
The supervised approaches use labeled samples to train the model. Early studies mostly employed supervised learning to classify the components. Nonetheless, early studies~\cite{Jindal2008,Mukherjee2013:} used datasets labeled as near-ground-truth or preferred to use a small set of manually labeled samples. More recent approaches focused on unsupervised and semi-supervised approaches to overcome several limitations such as the lack of labeled data, uncertainty of near-ground-truth datasets, etc. Supervised approaches employed either classical techniques or deep learning to perform the classification.

\subsubsection{Classical}
Early studies adopted classical classifiers such as the Naive Bayesian Model, SVM (Support Vector Machine), Multi-Layer Perceptron (MLP), and Logistic Regression (LR). Such classifiers utilized samples with ground-truth labels from human experts to train the detection models.

% Given the importance of the labels’ quality in fraud detection, the gold standard labeling for reviews was proposed by~\citet{Myle2012}. Myle \textit{et al.} incorporated a Bayesian model to create a standard gold labeling for fraud review detection. To classify reviews, SVM was used on six different datasets (including Priceline). 
% The results yielded an accuracy of 89.6\% on a dataset collected from Priceline. 
% are reported on different metrics; 89.1\% for Precision, 90.3\% for Recall, 89.7\% for F-score, and 89.6\% for Accuracy. 

% With the success of the Bayesian model in inferring the gold-standard labels, \citet{conf/acl/LiCL13} also proposed TopicSpam based on the Bayesian network. Li \textit{et al.} employed Latent Dirichlet Allocation (LDA) in TopicSpam to differentiate between fraud and genuine reviews based on the topic-word distribution. A Bayesian iterative algorithm was then applied to the trained latent variables to predict the final label of reviews. Evaluation on the TripAdvisor dataset showed an accuracy of 94\% for TopicSpam. 
% Early approaches also employed classical approaches to perform fraudster classification. 
Peng \textit{et al.}~\cite{Peng2014} proposed a relationship-based method to detect fraudsters. Hence, Peng \textit{et al.}~\cite{Peng2014} considered the reviews as a relationship between the item and the user in a network. The sentiment of each review was obtained as the first step in the feature extraction step. The sentiment score of the review was then compared to the rating of the review as the first indicator, called Interior Difference (ID). 

The second and third indicators were Rating Deviation (RD) and Sentiment Deviation (SD), respectively. The final fraudulent score of a user was calculated by a linear regression classifier. The experiments on the Reseller\_rating dataset showed an NDCG of 90\%. 

Previous studies~\cite{Myle2012,conf/acl/LiCL13,Peng2014} revealed the potential of classical approaches for \textit{multi-component} classification. Therefore, Yoo \textit{et al.}~\cite{Yoo2017} adopted the Loopy Belief Propagation (LBP) on a \textit{multi-component graph} consisting of users, items, and reviews. The algorithm stated different \textit{edge potential} assumptions on links between the components in the graph. Yoo \textit{et al.}~\cite{Yoo2017} then applied the assumptions in an iterative algorithm called “Supervised Belief Propagation" (SBP). SBP calculated the final probability of each review being a fraud. 

The performance of the SBP showed an AUC of 93\% on the Epinions dataset with 131,828 components (users, items, and reviews) and 841,372 edges.

To better utilize the potentials of the different classical supervised approaches, Kumar \textit{et al.}~\cite{Kumar2018} proposed \textit{hierarchical supervised learning} to detect fraudsters on the Yelp dataset. 
User-based features such as Review Gap (RG) and Review Count (RC) were extracted. The extracted features were then used as the input to different classical classifiers such as LR, K-Nearest Neighbor (KNN), NB, AdaBoost, RF, and SVM. LR showed an AUC of 72.3\% as the best classifier.

Classical supervised approaches showed a significant potential to handle fraud detection in different studies. However, such approaches suffer from a limitation in handling small datasets and require human experts to label the training dataset. Nevertheless, supervised learning has shown promising performance in fraud detection. 
% Fig.~\ref{fig:supervise} shows a timeline of studies that utilized classical supervised approaches.

\subsubsection{Deep Learning}
\label{sec:DL}
% Recent approaches have extensively employed deep learning for fraud review detection. Advances in big data, triggered the researchers to consider deep learning as a potential solution to perform complicated classification tasks such as fraud detection on social review platforms. 
Since 2016~\cite{Yafeng2016}, the fraud detection research community has witnessed the success of deep learning approaches. Mostly utilized as an end-to-end architecture, deep learning approaches take texts as the input for a prediction task of either classification or regression. In other words, deep learning mostly does not have explicit feature extraction, and instead combines feature representation learning and the classification step into one end-to-end network.\\

% Hence, Wang \textit{et al.}~\cite{Wang2018} proposed an approach based on an attention model to extract the important features of jointly learned behavioral and text-based features. Various indicators were extracted, such as the activity window and the percentage of positive reviews, as the behavioral features. Lingo-statistic text-based features such as bigram were also extracted alongside the behavioral features. Continuous Bag Of Word (CBOW) was calculated for each word and then fed to a CNN to extract a representation for each sentence. Then an attention-based representation was calculated and combined with behavioral and lingo-statistic text-based features. A softmax was used for the final classification. The proposed approach demonstrated an accuracy of 91\% on the Yelp dataset.

\begin{figure}
    \centering
    \includegraphics[width=\linewidth]{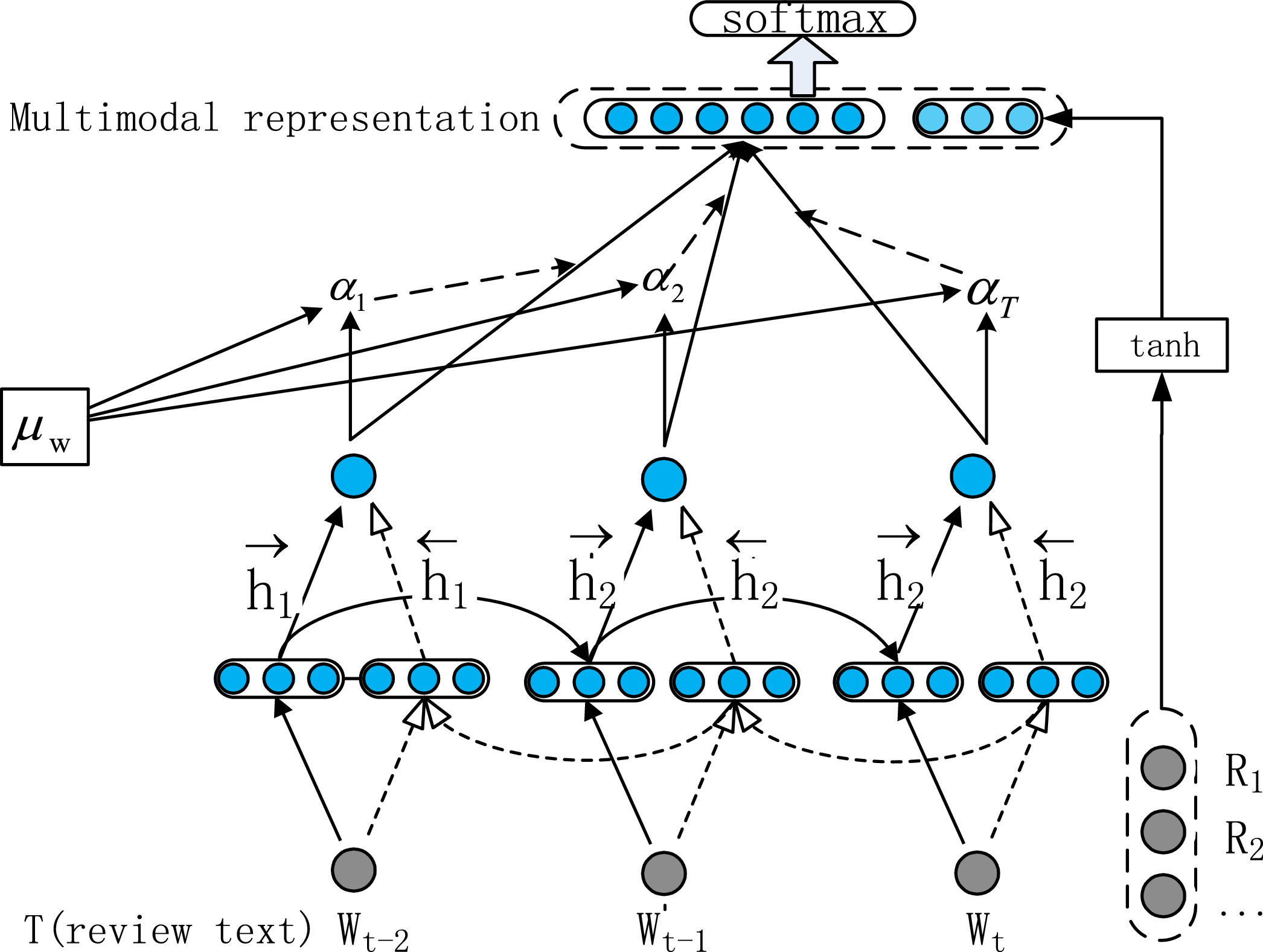}
    \caption{Learning multi-modal embedding representation of the n-gram textual and rich behavior features~\cite{liu2019opinion}.}
    \label{fig:liu2019opinion}
\end{figure}

% Recent studies also employed deep learning to address the limitations of the classical approaches. Previous classical approaches were scope limited and thus they lacked scalability. 
To address the domain dependent limitation of classical approaches, Liu \textit{et al.}~\cite{liu2019opinion} investigated traditional features for fraud review detection on a Chinese website. Liu \textit{et al.}~\cite{liu2019opinion} employed texts and metadata as multi-modal data to jointly learn a representation of each review. The architecture is given in Fig.~\ref{fig:liu2019opinion}. The joint representation was then fed to a bi-directional Gated Recurrent Unit (GRU) to learn the review text representation. Behavioral features such as the Ratio of Positive rating (PR), Burstiness (BST), and Rating Deviation (RD) were combined with the joint features to form a unique representation for each review. A CNN was then trained on these features to predict the class for each review. The dataset was collected from two large Chinese websites, namely Dianping hotels and restaurants. The results showed an F1-score of 68\% on the collected dataset. \\

\begin{figure}
    \centering
    \includegraphics[width=\linewidth]{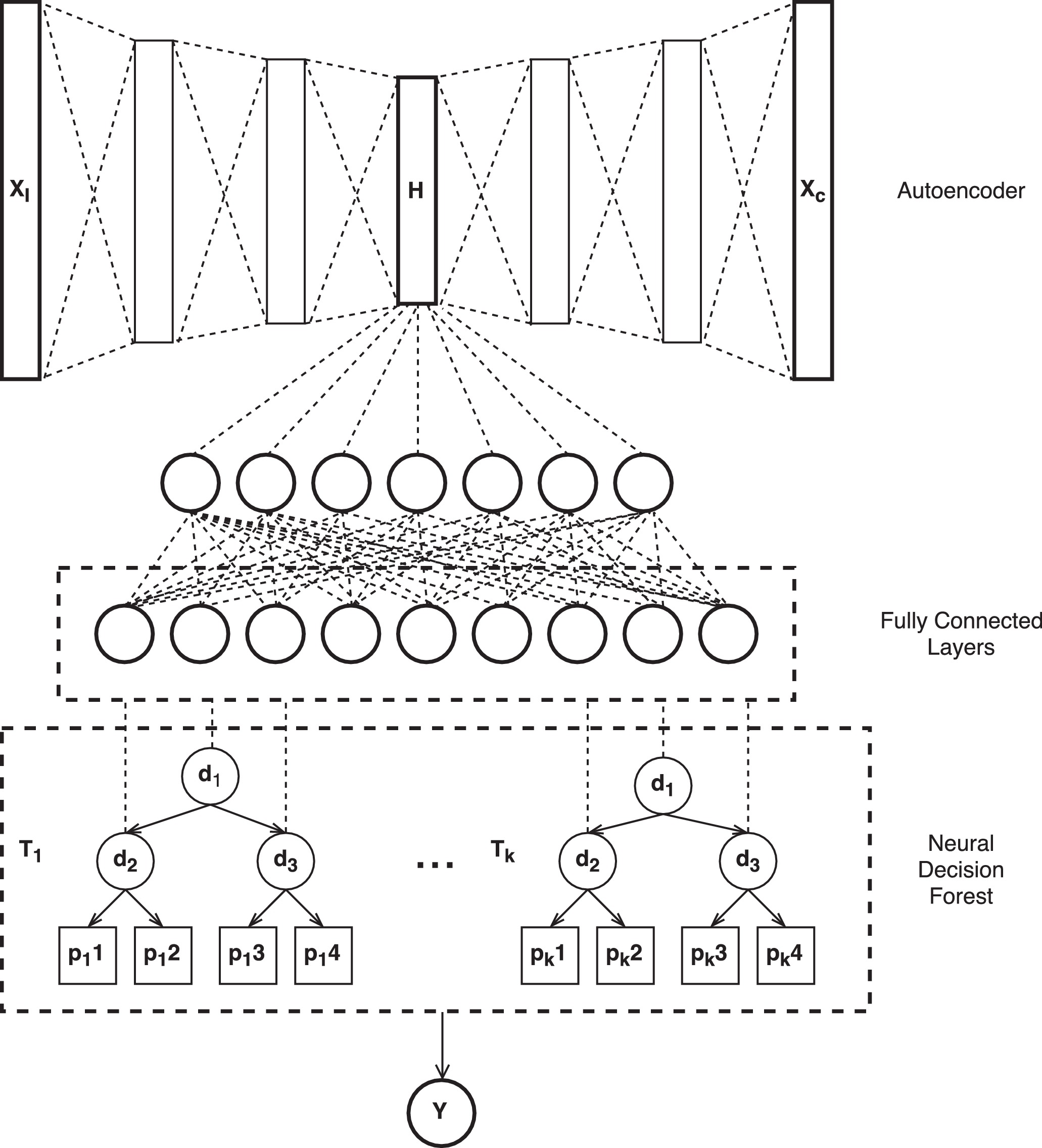}
    \caption{Proposed neural autoencoder decision forest in~\cite{dong2020opinion}.}
    \label{fig:Dong2020opinion}
\end{figure}

To utilize the potential of both classic and deep learning approaches, Dong \textit{et al.}~\cite{dong2020opinion} proposed a technique to combine deep learning and traditional classification. Both user’s behavioral data and text semantic features were extracted in the feature extraction stage. 
For the former, \textit{the entropy of scores}, \textit{review time entropy}, \textit{the entropy of ratings}, and \textit{the entropy of the product’s comment time} were extracted and combined with an indicator called the \textit{same date} indicator. For the latter, the review length and the review text semantics were extracted. 
An \textit{AutoEncoder} (AE) was applied to reduce the noisy components from the initial representation. An overview of AE is given in Fig.~\ref{fig:Dong2020opinion}. The bottleneck features were then used as the input to a \textit{Neural Decision Tree} (DT); i.e., a decision tree where each node represents a probabilistic distribution. The approach showed an accuracy of 95.85\% on the Amazon dataset. 

To compare the effectiveness of different deep learning approaches, Hajek \textit{et al.}\cite{hajek2020fake} proposed an approach to compare a Deep Feed Forward Neural Network (DFFNN) with a CNN. Review ratings and word embedding (learnt using skip-gram) were used as the features. Such features were then fed to a CNN and a DFFNN for the final classification. The CNN model achieved an accuracy of 81.30\% on the Amazon dataset.

Early studies~\cite{Yafeng2016,liu2019opinion} on deep learning mainly focused on performance improvement, recent studies~\cite{dong2020opinion,hajek2020fake} focused on different deep learning techniques to compare and analyze the potential of such approaches in the case of challenging fraud detection tasks. The state-of-the-art study~\cite{liu2019opinion} demonstrates the combination of deep learning with other aspects (e.g., multimodal data) resulted in a significant improvement. 
% Fig.~\ref{fig:DL} depicts a synthetic timeline of previous studies using deep learning for fraud detection.

\subsection{Semi-supervised}
\label{sec:semi-supervised}
Supervised approaches have the limitation of relying upon a quality labelled dataset by human experts which is expensive to obtain. Semi-supervised learning provides an opportunity to utilize the existing small set of labels to propagate the available knowledge to unlabelled samples. Different from supervised learning, there is no training set to train a model, but the model progressively learns the trend in data and tunes the hyper-parameters. 
% with an accuracy of random classifier~\cite{Shebuit2015}. 
% The recently proposed approaches mostly focused on using semi-supervised or unsupervised approaches to deal with the lack of labeled datasets. Semi-supervised approaches provide an opportunity to generalize the detection techniques to different domains. Hence, \citet{Wang2016} proposed a semi-supervised Recursive Autoencoder (RAE) to address the limitations of behavioral and traditional text-based features in capturing the global representation of a component. Word Embeddings (WEs) were extracted from the text using the Skip-gram model and then fed to an LR for the final classification. The proposed approach demonstrated an F1-score of 94\% on a Chinese dataset called Sina Weibo.\\
Given the effectiveness of deep learning in handling multimodal data, Deng \textit{et al.}~\cite{Deng2017} also proposed a framework with features extracted from both metadata and the content of the review text. Several features were extracted from the metadata as behavioral features:
\textit{User-Level (UL)} and \textit{User Mobility} from the IPs which were concatenated with the Bag of Word (BOW) as a text-based feature. The joint representation was then fed to an autoencoder for dimensionality reduction. K-Nearest-Neighbor (KNN) was applied to the features to achieve different feature clusters. A new dataset was collected from JD.com to evaluate the effectiveness of the proposed approach on different domains. The proposed approach showed an accuracy of 89.3\% on the JD dataset.
% Similarly, Narayan \textit{et al.} \cite{narayan2018review} proposed a PU-based approach on six different classifiers, and the best result was obtained with an accuracy of 78.12\% on Tripadvisor. 
% Rout \textit{et al.}~\cite{Rout2017} also proposed a semi-supervised approach using the PU learning to evaluate the effectiveness of the semi-supervised approach on different datasets. 
% Features such as sentiment polarity, Parts of Speech (POS) tags,  Word Count (WC), and bigram frequency counts were extracted from the text. The PU algorithm was applied to unlabelled reviews to provide the class probability through a label propagation algorithm. Next, an Expectation-Maximization (EM) was applied to the refined features (from the label propagation) as a semi-supervised approach. A scoring algorithm was also applied to the extracted features to provide the final scores. Several datasets were extracted using reviews from five different domains including TripAdvisor, Amazon, Yelp, Expedia, and Hotels.com. 
The best performance was obtained with an accuracy of 83.75\% on the TripAdvisor dataset.

Recently introduced semi-supervised approaches mostly utilized graph-based techniques to deal with fraud review. Shehnepoor \textit{et al.}~\cite{Shehnepoor2017} proposed NetSpam, combining text-based and behavioral features and obtaining the importance of each feature using a weighting method based on the \textit{Metapath} concept. The weights were used in a Heterogeneous Information Network (HIN) to link the reviews with similar weights. A novel scoring function was then proposed to calculate the probability of a review being a fraud. 
\begin{equation}
    Pr_{u,v} = 1 - \prod_{i=1}^L 1 - mp^{p_i}_{u,v} \times W_{p_i}
\end{equation}
where $Pr_{u,v}$ is the probability of user $u,v$ being fraudster based on the metapath value between two users on feature $p_i$ ($mp^{p_i}_{u,v}$) and $W_{p_i}$ is the weight of the feature. 
The results on the Yelp dataset demonstrated an AUC of 77\%. 

Given the effectiveness of the graph-based techniques, Yilmaz \textit{et al.}~\cite{Yilmaz2018} proposed a semi-supervised approach called SPR2EP. SPR2EP incorporated two types of features; \textit{review-based} features, and \textit{user-item} features. The former was extracted through an algorithm called Doc2Vec, and the latter was obtained through an algorithm called Node2Vec. Both approaches relied on Word2Vec to generate the embeddings. The proposed Doc2vec converted each document to a vector embedding with a size of 384. Node2Vec mapped the review text to the same vector embedding space of size 384. 
The concatenated features were then fed to a Linear Regression model for the final scoring. The best performance was obtained with an AUC of 83.18\% on the Yelp dataset.

As one of the latest studies, Wang \textit{et al.}~\cite{wang2022vote} proposed a vote-based integration model to harness the power of different heterogeneous information rank-based algorithms. The proposed framework, \textit{SpamVote}, took ranked lists generated by different unsupervised detection models as inputs, computed the weighting of ranked reviews based on the similarity between two ranked lists, and output the score of a review is a fraud review. \textit{SpamVote} showed NDCG@100 of 93.68\% on the Yelp dataset.
\subsection{Unsupervised}
\label{sec:unsupervised}
% Unsupervised approaches were developed to overcome the limitation of labeled datasets shortcomings. To address such a challenge, the unsupervised approaches potentially provide a great opportunity for researchers.  \\
% Early unsupervised techniques relied on proposing a better feature representation to deal with fraud reviews. \citet{Xu2015} proposed a Unified Review Spamming Model (URSM), utilizing LDA for the topic analysis. \citet{Xu2015} extended the model to a hierarchical LDA (hLDA). The hLDA model enabled the approach to create a  hierarchical tree of topics. Given the outputs of hLDA, \citet{Xu2015} claimed that the reviews with a general overview of the product is more probable to be a fraud review and vice-versa. The features extracted from the hLDA were then concatenated with the rating and burstiness to form the final representation of each review. A distribution was obtained for the review type (genuine/fraud) through an extension of the EM algorithm called Gibbs-EM. The performance of URSM was evaluated on three different datasets; Amazon, TripAdvisor, and Yelp. The best performance of URSM was reported on the Amazon dataset with an accuracy of 78.8\%.\\
In addition to their promising performance in semi-supervised models, graph-based approaches were also employed for unsupervised learning. 

Liu \textit{et al.} proposed HoloScope \cite{Liu:2017:HTA:3132847.3133018} that models data as a bipartite graph with users as the source nodes and items as the sink nodes. 
% % \citet{kumar2018rev2} 
% % \begin{itemize}
% %     \item \textbf{Component:} Users.
% %     \item \textbf{Challenges:} Cold start problem on different datasets. 
% %     \item \textbf{Approach:} Un-supervised. 
% %     \item \textbf{Feature:} Behavioral (review rating). 
% %     \item \textbf{Scorer:} Bayesian.
% %     \item \textbf{Dataset:} Amazon.
% %     \item \textbf{Accuracy:} 64.89\%.
% % \end{itemize}
% } ;

\begin{center}
\begin{table*}[hbt!]
% \captionsetup{font=8p}
\centering
\caption{Studies based on approaches.}\label{tab:approaches}
\begin{tabular}{|p{1.5cm}|p{1.5cm}|p{4cm}|p{3cm}|p{2cm}|}
\hline
Approach& Type&  Model & Output & Paper \\
\hline
\hline
\multirow{8}{*}{Supervised} & \multirow{4}{*}{Classical} & 
Support Vector Machine & Binary classes& Myle \textit{et al.}~\cite{Myle2012} \\ \cline{3-5}
& & 
Bayesian Network & Ranked scores& Li \textit{et al.}~\cite{conf/acl/LiCL13} \\ \cline{3-5}
& & 
Linear Regression & Binary classes & Peng \textit{et al.}~\cite{Peng2014} \\ \cline{3-5}
& & 
Linear Regression & Binary classes & Kumar \textit{et al.}~\cite{Kumar2018} \\ \cline{2-5}
& \multirow{4}{*}{Deep Learning} & Convolutional Neural Network & Binary classes& Wang \textit{et al.}\cite{Wang2018} \\ \cline{3-5}
&  & Convolutional Neural Network & Binary classes & Liu \textit{et al.}\cite{liu2019opinion} \\ \cline{3-5} 
&  & Neural Decision Tree & Binary classes & Dong \textit{et al.}~\cite{dong2020opinion}
 \\ \cline{3-5} 
&  & Convolutional Neural Network & Binary classes & Hajek \textit{et al.}~\cite{hajek2020fake}
 \\ \hline 
 \multirow{6}{*}{Semi supervised} & \multirow{5}{*}{Tabular} & 
Linear Regression & Binary classes& Wang \textit{et al.}~\cite{Wang2016} \\ \cline{3-5}
& & K-nearest Neighbor & Clusters & Deng \textit{et al.}~\cite{Deng2017} \\ \cline{3-5}
& & Linear Regression & Binary classes & Rout \textit{et al.}~\cite{Rout2017} \\ \cline{3-5}
& & Linear Regression & Binary classes & Yilmaz \textit{et al.}~\cite{Yilmaz2018}  \\ \cline{3-5} 
& & Voting & Binary classes & Wang \textit{et al.}~\cite{wang2022vote}   \\ 
\cline{2-5}
& Graph & Heterogeneous Information Network & Ranked scores & Shehnepoor \textit{et al.}~\cite{Shehnepoor2017} \\ \hline
 \multirow{5}{*}{Unsupervised} & \multirow{3}{*}{Tabular} & 
Expectation Maximization & Ranked probabilities & Xu \textit{et al.}~\cite{Xu2015}\\ \cline{3-5}
& & Maximum Likelihood Estimation & Binary classes & Dong \textit{et al.}~\cite{dong2018unsupervised}\\ \cline{3-5}
& & Ranking & Ranked scores & Xu \textit{et al.}~\cite{Xu2019} \\ \cline{2-5}
& \multirow{2}{*}{Graph} & Iterative Mutually Exclusive & Ranked scores &  Hooi \textit{et al.}~\cite{Hooi:2017:GFD:3119906.3056563}\\ \cline{3-5}
& & Bayesian network & Ranked scores & Kumar \textit{et al.}~\cite{kumar2018rev2} \\ \hline

\end{tabular}
\end{table*}
\end{center}

Nodes were linked through a directed edge signed by either rating or the timestamp of the review. Holoscope then applied a mutually exclusive iterative algorithm inspired by~\cite{Chengzhang2019Network,Hooi:2017:GFD:3119906.3056563} to score the users as fraudsters or genuine. The results on the Yelp dataset demonstrated an AUC of 99.5\%. 

% Topic sentiment models were also broadly used in the unsupervised approaches through LDA on fraud detection. An Unsupervised Topic-Sentiment Joint (UTSJ) probabilistic model was proposed by Dong \textit{et al.}~\cite{dong2018unsupervised} to employ topic modeling for the fraud detection task. The UTSJ assumes a high correlation between the sentiment and the review topic. Hence, an LDA is employed to model the topic as the sentiment indicator of the reviews. A Maximum Likelihood Estimation (MLE) was then applied to the extracted topic sentiment features to obtain the fraudulent scores of the reviews. The USTJ showed an F-score of 83.53\% on the Yelp dataset.\\
Hooi \textit{et al.}~\cite{Hooi:2017:GFD:3119906.3056563} also proposed a novel unsupervised approach to deal with a recent hot-topic challenge in fraud detection: the \textit{camouflaged fraudsters}. \textit{Camouflage} refers to an act of writing genuine reviews by fraudsters to escape detection. Hooi \textit{et al.} mapped the components to a bipartite network with items and reviews as two different node types in the graph. An iterative algorithm, inspired by~\cite{kumar2018rev2}, was applied to a bipartite network to find the score of a new user potentially being a fraudster. Results on Amazon showed an accuracy of 89\%. 
Recently, unsupervised approaches are also employed to address new challenging topics such as the cold-start problem. Kumar \textit{et al.}~\cite{kumar2018rev2} proposed REV2 to incorporate a Bayesian Belief Network to perform inference on three components (fairness of the user, the goodness of the product, and the reliability of a reviewer) from five different datasets including Flipkart, Bitcoin OTC, Bitcoin Alpha, Epinions, and Amazon. To handle the cold-start problem, Kumar \textit{et al.} utilized a Laplacian smoothing. To score a user as a fraudster or honest, a mutually recursive algorithm was applied to the components. Results demonstrated an accuracy of 64.89\% for the multi-component classification on the Amazon dataset.

As unsupervised approaches are not limited by the labeling quality of the reviews, recent research mostly relied on unsupervised methods to address new challenging topics (such as cold start and camouflage). The recent approaches mainly employed graph-based techniques to model the components. The potential of unsupervised approaches is not only to improve the detection accuracy in fraud detection but also to address the current and future challenges. 
% Fig.~\ref{fig:unsupervised} shows a timeline of the most recent studies on unsupervised approaches. 

\subsection{Summary}
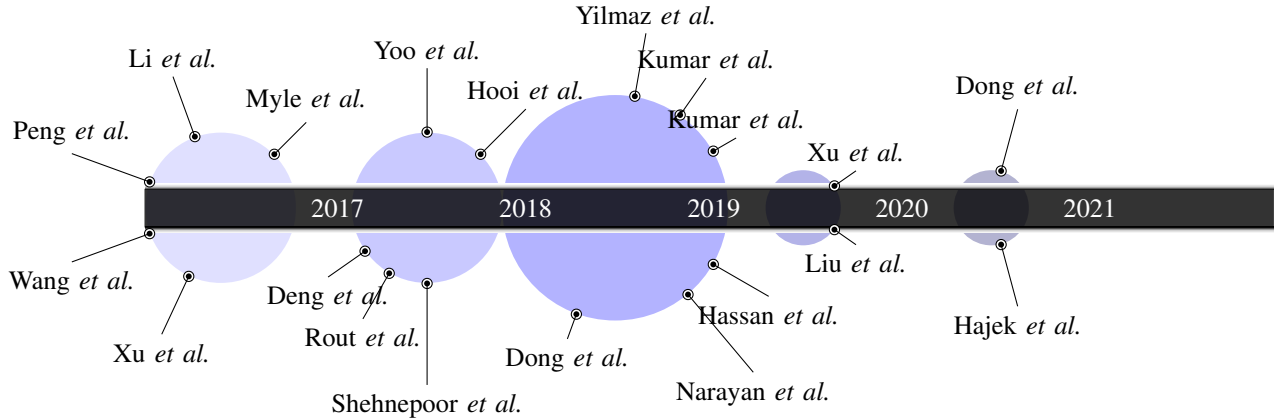
\begin{figure*}
    \centering
    \begin{tikzpicture}[scale = 0.5,timespan={}]
    \timeline[custom interval=true]{2017,...,2021}     
    \begin{phases}
        \initialphase{involvement degree=2cm,phase color=blue!40!white}%5
        \phase{between week=1 and 2 in 0.5,
          involvement degree=2cm,phase color=blue!70!white}%5
        \phase{between week=2 and 3 in 0.5,
          involvement degree=3cm,phase color=blue!100!white}%7
        \phase{between week=3 and 4 in 0.5,
          involvement degree=1cm,phase color=blue!70!black}%2
        \phase{between week=4 and 5 in 0.5,
          involvement degree=1cm,phase color=blue!40!black}%2
      \end{phases}
    
    % % 2016
      \addmilestone{at=phase-0.45,direction=45:1.2cm,
        text={Myle \textit{et al.}}, text
        options={above}}
      \addmilestone{at=phase-0.110,direction=110:1.7cm,
        text={Li \textit{et al.}}, text
        options={above}}
      \addmilestone{at=phase-0.160,direction=160:2.2cm,
        text={Peng \textit{et al.}}, text
        options={above}}
      \addmilestone{at=phase-0.200,direction=200:2.2cm,
        text={Wang \textit{et al.}}, text 
        options={below}}
      \addmilestone{at=phase-0.245,direction=245:1.7cm,
         text={Xu \textit{et al.}}, text options={below}}
    % % 2017
      \addmilestone{at=phase-1.90,direction=90:1.7cm,
        text={Yoo \textit{et al.}}, text 
        options={above}}
      \addmilestone{at=phase-1.45,direction=45:1.7cm,
        text={Hooi \textit{et al.}}, text 
        options={above}}
      \addmilestone{at=phase-1.270,direction=270:2.7cm,
        text={Shehnepoor  \textit{et al.}}, text 
        options={below}}
      \addmilestone{at=phase-1.215,direction=215:1.2cm,
        text={Deng  \textit{et al.}}, text 
        options={below}}
      \addmilestone{at=phase-1.240,direction=240:1.4cm,
        text={Rout  \textit{et al.}}, text 
        options={below}}

    % % 2018
      \addmilestone{at=phase-2.80,direction=80:1.7cm,
        text={Yilmaz \textit{et al.}}, text
        options={above}
        }
      \addmilestone{at=phase-2.55,direction=55:1.2cm,
        text={Kumar \textit{et al.}}, text
        options={above}
        }
      \addmilestone{at=phase-2.30,direction=30:0.7cm,
        text={Kumar \textit{et al.}}, text
        options={above}
        }
      \addmilestone{at=phase-2.250,direction=250:0.7cm,
        text={Dong \textit{et al.}}, text
        options={below}
        }
      \addmilestone{at=phase-2.310,direction=310:2.7cm,
        text={Narayan \textit{et al.}}, text
        options={below}
        }
      \addmilestone{at=phase-2.330,direction=330:1.7cm,
        text={Hassan \textit{et al.}}, text
        options={below}
        }
        
    % % 2019
      \addmilestone{at=phase-3.35,direction=35:0.7cm,
        text={Xu \textit{et al.}}, text
        options={above}
        }
      \addmilestone{at=phase-3.325,direction=325:0.7cm,
        text={Liu \textit{et al.}}, text
        options={below}
        }

    % % 2020
      \addmilestone{at=phase-4.75,direction=75:1.7cm,
        text={Dong \textit{et al.}}, text
        options={above}
        }
      \addmilestone{at=phase-4.285,direction=285:1.7cm,
        text={Hajek \textit{et al.}}, text
        options={below}
        }

    \end{tikzpicture}
    \caption{The volume timeline of the most recent studies using different approaches for fraud detection. Diameter of each circle is proportional to the number of studies for one year period [2017-2018, 2018-2019, etc.]}
    \label{fig:approach-tl}
\end{figure*}

The volume timeline of previous studies on different approaches is provided in Fig.~\ref{fig:approach-tl}. Similar to Fig. \ref{fig:features-tl}, there is a boost in the number of studies focusing on models as the main contribution in 2018, coupled with the introduction of deep learning for fraud detection. Thereafter, most studies concentrated on deep learning for the feature representation learning step. A summary of approaches is given in Fig.~\ref{tab:approaches}.

\section{Analysis}
\label{sec:discussions}
This section includes a detailed analysis of the findings in various subsections. We provide an analysis of three types of topics: \textit{classical}, \textit{ongoing}, and \textit{future} challenge. In the first section, we provide an analysis of classical topics already investigated in previous studies. After that, we provide an overview of possible future directions. In the subsequent section, we examine data as the primary challenge, in previous and future studies. We also suggest future steps to address the shortage of data. Finally, we discuss different future topics and possible solutions.

% Then, an analysis of the recent studies and an overview of possible future topics for each step of the proposed synthetic framework are provided.

\subsection{Discussion on Topics: Classical Challenging Topics}
\label{sec:topic-anal}
% {\color{blue}
% As we provided a comprehensive study on different topics on fraud detection, in this subsection first we give a deep discussion in the first section called as Discussion. In the next subsection, the findings of the current study on different topics are provided. 
% }
% \subsubsection{Discussion}

\begin{figure}
    \centering
    \pgfplotstableread{
    % Label1 Component     
    % Review 55.38    
    % User 38.61    
    % Item 6.16  
    % Label2 Feature
    % Joint 38.61
    % Behavioral 29.82
    % Text-based 31.57
    % Label3 Approach
    % Supervised 45
    % Semi-supervised  31.66
    % Unsupervised 23.34
    Label 2 3 4    
    Component(\%) 55.38 38.46 6.16
    Feature(\%) 38.61 29.82 31.57    
    Model(\%) 45 31.66 23.34  
    }\testdata

    \begin{tikzpicture}

    \begin{axis}[
        ybar stacked,
        ymin=0,
        ymax=108,
        xtick=data,
        legend style={at={(0.5,-0.1)},anchor=north}, 
        % legend pos=outer north east,
        reverse legend=true, % set to false to get correct display, but I'd like to have this true
        xticklabels from table={\testdata}{Label},
        xticklabel style={text width=2cm,align=center},
    ]
    \addlegendentry{Review (C), Joint (F), Supervised (M)}
    \addplot [fill=red!60] table [y=2, meta=Label, x expr=\coordindex] {\testdata};
    \addlegendentry{User (C), Behavioral (F), Semi-supervised (M)}
    \addplot [fill=green!80] table [y=3, meta=Label, x expr=\coordindex] {\testdata};
    \addlegendentry{Item (C), Text-based (F), Unsupervised (M)}
    \addplot [fill=blue!60,nodes near coords,point meta=y] table [y=4, meta=Label, x expr=\coordindex] {\testdata};
    % \addplot [
    %     ybar, % this makes it show the total for some reason
    %     nodes near coords,
    %     nodes near coords style={%
    %         anchor=south,%
    %     },
    % ] table [ y expr=0.00001, x expr=\coordindex] {\testdata};

    \end{axis}
    \end{tikzpicture}    
    % \begin{tikzpicture}
    % \tikzset{lines/.style={draw=white},}
    % \pie[cloud,text=inside,color={red!20,red!40,red!60},sum=auto, after number={\%},every only number node/.style={text=black},style={lines}]{55.38/Review,38.46/User,6.16/Item}
    % \node[below=23mm of O] {\textbf{Components}};
    % \pie[pos={8,0},cloud,text = inside,color={green!20, green!40, green!60},sum=auto, after number={\%},every only number node/.style={text=black},style={lines}]{38.61/Joint,29.82/Text-based,31.57/Behavioral}
    % \node[below=23mm of O] {\textbf{Features}};
    % \pie[pos={4,-8},cloud,text = inside,color={ blue!20,blue!40, blue!60},sum=auto, after number={\%},every only number node/.style={text=black},style={lines}]{45/Supervised,31.66/Semi-supervised,23.34/Unsupervised}
    % \node[below=23mm of O] {\textbf{Approaches}};
    % \end{tikzpicture}
    \caption{Percentage breakdown of studies on components, features and classification approaches.}
    \label{fig:total-analysis}
\end{figure}
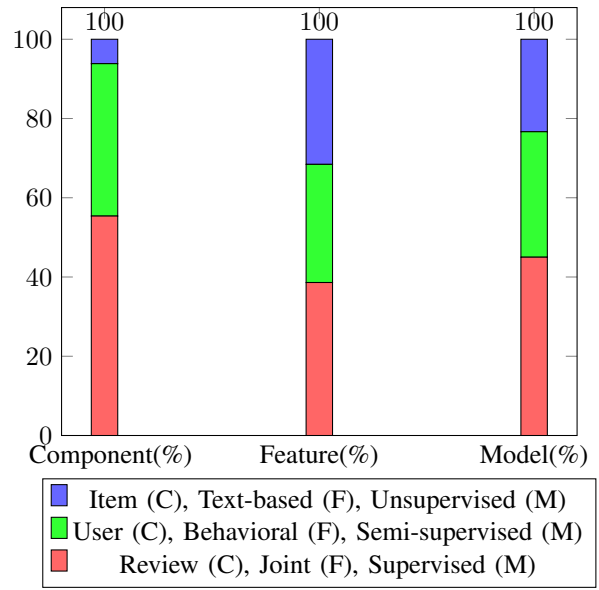

This survey studied 58 recent works with 80\% after 2016 (46 out of 58) to properly present the most recent techniques and challenges in fraud detection. Fig.~\ref{fig:total-analysis} shows the research study statistics for different topics: components, features, and approaches.
%Obviously, the stats on fields of study (detection, generation) and the classifier are not depicted, since the studies are either based on one category or too diverse. 
As Fig.~\ref{fig:total-analysis} suggests, user and review detection were the main focus of researchers in recent years (94\% combined), while item (as a separate component) takes up about 6\% of the research efforts. 

\subsubsection{Targeted Item Detection}
% All studies were, in combination with the other components, in the form of a multi-component classification. However,
Target item detection benefits fraud detection in different aspects. For example, it facilitates fraudster group detection, as suggested by Ji \textit{et al.}~\cite{ji2020burst}. It has also shown to be effective in dealing with the cold-start problem~\cite{wang2017handling,you2018attribute} by finding items with a burst of reviews written in a short period of time. This encourages researchers to prioritize target item detection in order to address potential future challenges. \\
\subsubsection{Feature Extraction Importance}
The uniform distribution over the studies on fraud features (Fig. \ref{fig:total-analysis}) suggests an equal importance of each category for researchers. It is worth mentioning that early studies employed text-based features to address fraud detection. In recent years, and with the significant advances in deep neural networks, deep learning techniques were utilized to represent text-based features implicitly using word embeddings. Deep learning was also utilized as a classifier in the classification step or in the joint learning task in end-to-end neural networks. End-to-end approaches provide an opportunity to address fraud detection, by combining the feature representation and classification step. Despite behavioral features’ effectiveness in handling fraud review, in recent years, and with the growth of bot-generated reviews, the behavioral features' representation may result in an unreliable representation of each component. However, given that new datasets contain information of the lower layers in the OSI model (such as IP, MAC address, and other basic information of users), these will likely compensate for the shortcomings in the application layer metadata and review text. 
% Another way to deal with suspicious behavior is through Graph Convolutional Networks (GCNs). GCNs are among the recent approaches employed for different modeling tasks of fraud detection. However, GCNs are overlooked in different applications, such as the behavior modeling of reviewers in a group, affected by co-reviewers in a group. GCN~\cite{kipf2016semi} aims to encode both the local graph structure and the nodes attribute to obtain a refined representation of each node in a graph. The refined representation $Z$ for each node is then obtained through the function $f(X, A)$:
% \begin{equation}
% \label{eq:GCN-softmax}
% f(X,A) = softmax(\hat{A}ReLU(\hat{A}XW^{(0)})W^{(1)})
% \end{equation}
% where $X$ is the input of the network. In Eq. \ref{eq:GCN-softmax} $\hat{A} = A + I_N$, where the matrix $A$ represents the adjacency matrix of the graph and the $I_N$ is the identity matrix to include the self-connections in the network. $W^{(0)}$, $W^{(1)}$ represent the weights of the network in the first and second layer, respectively. After applying the $ReLU$ function in the first layer, softmax is applied in the second layer for the final representation. A GCN has a potential to be used for fraudster detection tasks, where similarities between representations play the most important role.\\
Joint feature learning is considered to be an active area of research, since advances in both text-based and behavioral features result in improved joint representations.
\subsubsection{Graph-based Approach Importance}
According to Fig. \ref{fig:total-analysis}, almost half of the approaches adopted supervised learning, one-third of the approaches utilized semi-supervised learning and the rest used unsupervised learning. 

Fig.~\ref{fig:total-analysis} suggests that there is a tendency towards using un/semi-supervised approaches. More than half of the studies relied on unlabelled data (unsupervised and semi-supervised approaches combined), while the majority of the remaining approaches, with supervised learning as their main focus, relied on the Yelp dataset.

With the lack of data and explanations provided in previous sections, a set of  suitable approaches with a capability to correctly populate the labels for the unlabelled samples is required. Graph-based approaches, however, empower the models to utilize the merits of the connection between components to better generalize the classifications. Conclusively, the findings for the topic analysis demonstrate that graph-based approaches, as the latest techniques to address the lack of labels, can be utilized to employ a message forwarding~\cite{Akoglu2013} or inductive learning~\cite{Hooi:2017:GFD:3119906.3056563} to populate the labels based on different criteria.  However, such propagation approaches (message forwarding, etc.) are well explored (Fig.~\ref{fig:total-analysis}) and researchers are using different refinement methods to deal with the representation tuning.

\subsection{Lack of Data Problem: Ongoing Challenging Topics}
\label{sec:data}
% {\color{blue} Lack of data, as the most important challenge in fraud detection is discussed in following subsections. First, we provide a discussion on data challenges and then we explore the main findings of the current study.}
% \subsubsection{Discussion}
The lack of data is the most challenging topic for researchers in fraud detection. The challenges are across a range of different tasks across all stages in our systematic fraud detection framework, such as the data collection for a component, data preparation for feature extraction, and most importantly the labels used to train or evaluate the framework performance for the classification/scoring step.
% Most studies in fraud detection discuss the data challenges in terms of collection, organizing, and most importantly the labeling. 
\subsubsection{Labels}
The most pressing challenge is the lack of trustworthy ground truth for fraud reviews, due to the challenging nature of fraud labeling.
% since determining whether a review is a fraud or a genuine one is a challenging task, 
Due to the complexity of the task, even human labeling accuracy is no better than a random classifier~\cite{Shebuit2015,Yuanshun2017}. Some platforms (e.g., TripAdvisor) employed review experts as human judges to label the reviews~\cite{Aghakhani2018}. Furthermore, the difficulty of labeling depends on the task. For example, a group labeling task is less challenging than labeling individual users, since the collaboration among members of a group provides a reasonable context for comparison and judgement~\cite{ji2020burst}.
Similarly, some platforms use detection algorithms to label the datasets, e.g., Yelp~\cite{Bitarfan2019,Zhang2020gcn,Yilmaz2018,Wang2018,wang2017handling,you2018attribute,Li2019,Luca2016}. Datasets labeled by the detectors, are called ``near ground-truth" datasets, and the labels are mostly used in semi-supervised or unsupervised approaches~\cite{Shebuit2015,Shehnepoor2017}. Apart from review labeling challenges, the other challenge is the lack of ground truth for the other components, such as users (fraudster or genuine user) and items (targeted by fraudsters or non-targeted). Some studies label the users with at least one fraud review as fraudsters~\cite{Shebuit2015}. Such labeling would result in misclassification, given that the labels are provided as near-ground truth.

% One solution can be to use a probability-based labeling procedure based on the number of fraud reviews written by the user. 
% Besides, the increasing trends towards different challenging areas such as user group detection \cite{zhang2020label,Xu2019,Zhu2019}, cold-start  \cite{wang2017handling,you2018attribute}, an urgent need to targeted item identification is likely recognized.
There is a growing trend towards using datasets populated by web crawling applications, to deal with the lack of data problems. The Tencent dataset has been extensively used in recent studies~\cite{Wang2019FdGars,Wen2020}, and provided labels as near-ground truth. 
Fig. \ref{fig:dataset-dist} shows that a considerable number of studies use the Yelp dataset. The first possible explanation is that the Yelp platform is a public dataset and is accessible for different purposes. The Yelp dataset also provides the ground truth for reviews based on the platform recommender. Similarly, the Amazon dataset provides a variety of datasets for different businesses with near-ground truth labels. TripAdvisor relies on a balanced distribution of fraud and genuine reviews for hotels in Chicago, providing 800 reviews for each class of fraud and genuine reviews. Such a dataset is either used in studies with classic classifiers~\cite{Rout2017,narayan2018review,Hassan2019} or for data augmentation to overcome data paucity ~\cite{Aghakhani2018,shehnepoor20}. The remaining datasets are crawled using web crawlers, mainly to provide an insight for different fraud detection topics.

% \subsubsection{Findings}
% With different datasets description and analysis provided in previous sections, numerous findings and research directions are discussed in following.\\  
Labeling the review texts requires different levels of knowledge. Similar studies in related areas might be interesting and inspiring in fraud detection. Studies on detecting fake news, as another type of spam, employed fact-checking to improve the accuracy of the labeling~\cite{fake-news-pathak2019breaking,fake-news-rashkin2017truth,fake-news-vijjali2020two}. Hierarchical fact checking is employed to find the clues about the origin of the news. However, accessing that level of information requires specific permission to the data. Nonetheless, fake news detection accuracy somehow depends on different datasets presented to the algorithm. In fraud review detection, on the other hand, acquiring such a relationship between different reviews in multiple platforms is challenging. Fake news detection, however, can provide some useful insights into how hierarchical fact checking~\cite{WANG2022107910} can be helpful to provide a degree of confidence on the labeling. Such an exploration requires considerable efforts on the labeller end.  

\textbf{As another direction,} incorporating expertise from other areas such as psychology could be useful in identifying key implications of written reviews and the intention of the user through the use of psychological indicators. Such indicators can be driven from the sentiment analysis of the review text. The indicators are then correlated with the analysis from a psychological point of view~\cite{psychology-porshnev2013machine,psychology-li2020impact}. Porshnev \textit{et al.}~\cite{psychology-porshnev2013machine} studied the emotional state of different users in Twitter using sentiment analysis. Word frequency is also used as a feature to analyze the psychological state of the people who experienced a sudden change in the isolation with the COVID-19 outbreak. The outcome of such psychological indications can be used to help the labeller. 

Given all the explanations and directions, achieving a large amount of data requires considerable investment from companies. Marciano \textit{et al.}~\cite{big-company-152} reported that fraud reviews in online shops cost \$152 billion a year. Facebook introduced new machine learning techniques to address scam on Facebook in 2018~\cite{big-company-facebook}. Although such contents are considered as opinion spam, finding the accurate label for the task requires significant investment to provide both budget and data to empower the labeling tools for finding the correct labels. 

\subsubsection{Multi-modal Data}
Fortunately, several platforms (e.g., Yelp, Amazon, and TripAdvisor) provide datasets including the review text and the metadata (e.g., user ID, item ID, review ID, date of a written review, rating given by the review, and the label) for each review. 
% However, the platforms provide limited information for datasets to protect their intellectual property and also to keep the platform secure.
However, the platforms also need to withhold important information to secure the users' privacy.
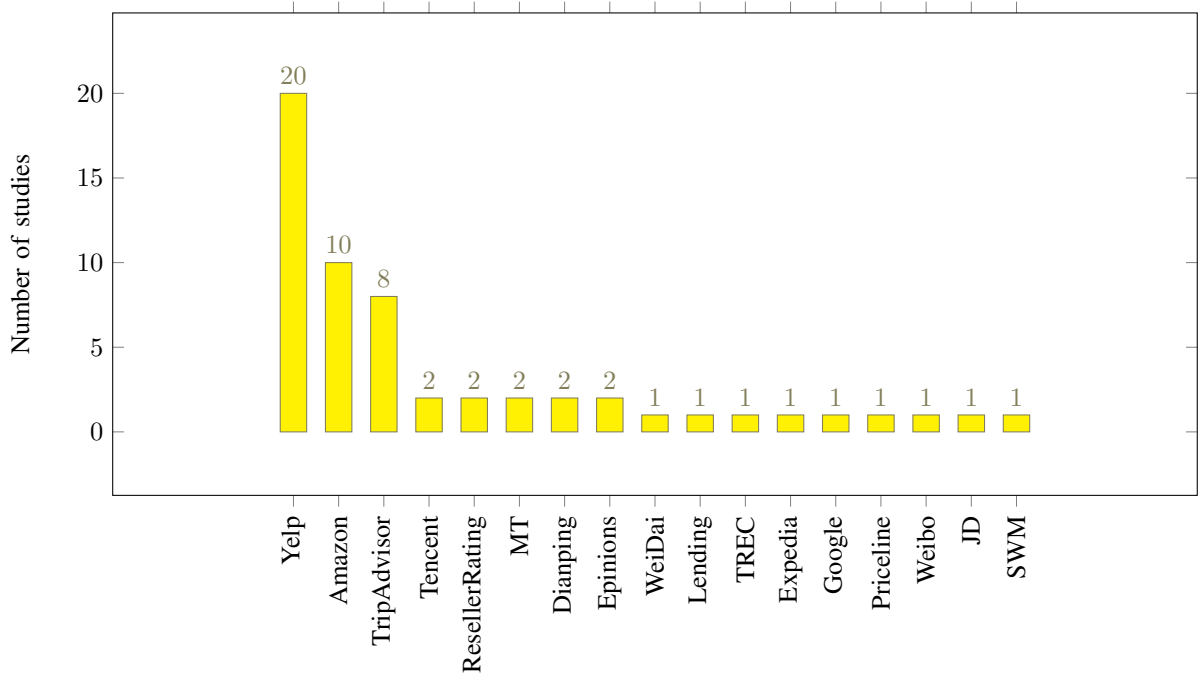
\begin{figure*}
    \centering
\tikzexternaldisable
\begin{tikzpicture}
\pgfplotsset{%
    width=16cm,
    height = 8cm
}
\begin{axis}[
    ybar,
    enlargelimits=0.25,
    % legend style={at={(0.5,-0.15)},
    %   anchor=north,legend columns=-1},
    ylabel={Number of studies},
    x tick label style = {rotate=90,anchor=east, align=left},
    symbolic x coords={Yelp,Amazon,TripAdvisor,Tencent,ResellerRating,MT, Dianping,Epinions,WeiDai,Lending,TREC, Expedia,Google,Priceline,Weibo,JD,SWM},
    xtick=data,
    xticklabels={
    Yelp,
    % ~\cite{dfraud},
    Amazon,
    % ~\cite{ji2020burst},
    TripAdvisor,
    % ~\cite{Aghakhani2018},
    Tencent,
    % ~\cite{Wang2019FdGars},
    ResellerRating,
    % ~\cite{wang2011review},
    MT,
    % ~\cite{crawford2016reducing},
    Dianping,
    % ~\cite{Li2016},
    Epinions,
    % ~\cite{Yoo2017},
    WeiDai,
    % ~\cite{He2020},
    Lending,
    % ~\cite{Zhao2019},
    TREC,
    % ~\cite{Lai2011}, 
    Expedia,
    % ~\cite{Banerjee2015},
    Google,
    % ~\cite{Hernandez2018},
    Priceline,
    % ~\cite{Myle2012},
    Weibo,
    % ~\cite{Wang2016},
    JD,
    % ~\cite{Deng2017},
    SWM,
    % ~\cite{Akoglu2013}
    },
    nodes near coords,
    nodes near coords align={vertical},
    ]
\addplot[yellow!40!black, fill = yellow!100!black] coordinates {(Yelp,20) (Amazon,10) (TripAdvisor,8) (Tencent,2) (ResellerRating,2) (MT,2) (Dianping,2) (Epinions,2) (WeiDai,1) (Lending,1) (TREC,1) (Expedia,1) (Google,1) (Priceline,1) (Weibo,1) (JD,1) (SWM,1)};
%\legend{used,understood,not understood}
\end{axis}
\end{tikzpicture}
\tikzexternalenable
    \caption{Dataset distribution over different studies. Note that the most recent study on each dataset is provided alongside the dataset.}
    \label{fig:dataset-dist}
\end{figure*}

% due to the reasons such as trade secrecy or continuous dynamic changes in platforms' databases.

3)	More specifically, Yelp, which has been the most studied dataset (as shown in Fig.~\ref{fig:dataset-dist}) includes the necessary metadata such as user ID, item ID, date of the written review, and the rating given by the review that can be used to extract behavioral features for user activity at the application layer of the OSI model~\cite{day1983osi}. However, experienced fraudsters employ a camouflage strategy to simply cover up their behavioral footprints and manage their traceable metadata. Although different approaches were proposed to address the camouflage problem, the performance of the state-of-the-art approaches is limited due to the lack of information of the user behavior at the lower layer of the OSI model (e.g., network layer). To deal with the camouflage problem more information is required to help the detection algorithm to fully trace the user’s behavior. Unfortunately, the well-known platforms are only providing particular confidential information such as the user IPs in each session, location coordinates, etc. It is worth mentioning that some platforms provide privileges (e.g. APIs) for the crawlers to collect data from different data modalities, also referred to as \textit{Multi-modal} data~\cite{chiu2017uncovering}. 
Gathering information from different modalities for feature extraction is an unexplored and a potential topic to consider for future studies~\cite{chiu2017uncovering}. Similarly, various studies employed multi-modal data to address spam tweets~\cite{jiang2015general,jiang2016spotting} on different domains of the social review platforms. 
% Similar to fraud review detection, tweet spam detection utilize burstiness of reviews to determine fraudsters and fraud reviews. 
Using different data models, e.g., the Denial of Service (DoS) containing different attributes, is an effective approach in detecting fraud reviews. Liu \textit{et al.}~\cite{liu2019opinion} suggest that feature types can be regarded as different data modalities. The features are likely extracted from one dataset with the metadata as the attribute to then extract text-based and behavioral features as different modalities.  Attributes from the network layer (e.g., IP of the user) were utilized in a few studies~\cite{Wang2019FdGars,Wen2020} to initialize a new area of research in fraud detection. Hence, future studies can extract multi-modal data and combine such features to achieve a better behavioral representation of each user. 

For example, Wen \textit{et al.}~\cite{Wen2020} and \textit{Wang et al.}~\cite{Wang2019FdGars} collected data from the Tencent platform which includes IP and different physical layers alongside information from the application layer. Technically speaking, such multimodal data enables the detection model to learn a joint representation and then extract a unique vector for each user. The vector not only represents the traceable user's behavior at the application layer but also the background activities of the users. Such additional information help the detector with new fraudster detection in a shorter time compared to a situation in where the detector is provided with the information from only one data modality, and this can also address the cold-start problem~\cite{wang2017handling}.

However, the inconsistency between the different modalities of data could also be an interesting topic to explore. For example, different studies considered rating as a sentiment indicator, representing the tendency of the user towards the item. Surprisingly, with the sentiment analysis of a text, such assumptions are more likely to be inaccurate~\cite{wang2017handling,Akoglu2013}, due to the inconsistency between the rating and the real sentiment of the review. Hence, to obtain accurate semantics, studies should solely rely on the sentiment analysis of the review text (and not rating). To train a joint representation of the text and metadata, several objective functions are proposed in different studies~\cite{wang2017handling,shehnepoor20}.

% Such representation is more likely to reveal the background activities of fraudsters. 

\subsubsection{Transfer Learning}
\textbf{One of the most recent topics} is to use transfer learning to improve fraud detection accuracy. Gupta \textit{et al.}~\cite{TL-gupta2021leveraging} employed BERT and transferred an NLP task to fraud review detection with different configurations. However, the study employed transfer learning as part of a representation extraction step from the text, though transfer learning can be applied to different steps of a machine learning task. To be specific, in a relatively similar area, such as speech recognition, studies applied transfer learning to improve the accuracy of the children's speech recognition using a model already trained with adult data~\cite{TL-matassoni2018non,TL-shivakumar2020transfer}. The transfer learning approach has shown to be useful in the case of data scarcity, as with a model trained with plentiful data on the other task, replacing the classification layer and initial training of the layer which requires a relatively smaller set of data~\cite{TL-shen2020deep}.

\subsection{Hot Topics: Future Challenging Topics}
\label{sec:future-works}
% {\color{blue} The future works of fraud detection depends on the analysis we provided in previous sections. In this sections, similar to previous ones, first we provide a discussion on future works and the findings are discussed in the next subsection.}
% \subsubsection{Discussion}
% Given the proposed systematic framework, we examine the most challenging topics for each of the three stages: component selection, feature extraction, and classification/scoring in fraud detection. 
There are some recent challenging topics addressed only by a few studies. We discuss such topics in the following and provide insights on possible solutions.
\subsubsection{Bot Generated Review Detection}
Bot review generation was introduced by Yuanshun \textit{et al.}~\cite{Yuanshun2017} and then explored by different studies \cite{Aghakhani2018,shehnepoor20}. Yuanshun \textit{et al.}~\cite{Yuanshun2017} claimed that bot-generated reviews are indistinguishable from human-written reviews from different aspects. Such reviews are manipulated by fraudsters, and generated in large volume, with minimum cost. A generator can be trained to generate authentic reviews, leaving neither human traces nor footprints. Consequently, behavioral features become less useful for bot-generated review detection. 
Yao \textit{et al.}~\cite{yao2017automated} reported an important observation on a bot trained by a character-based language model over the whole corpus. The study discussed one way to detect such fraud reviews is to parameterize the character distribution, as the distribution is one key factor in determining a bot review. Relying on a specific distribution, the study detected the bot-generated review using a likelihood measurement. This can be applied to different platforms (Yelp, TripAdvisor, etc.) using different criteria, as each domain has its own specific terminology. 

Shehnepoor \textit{et al.}~\cite{shehnepoor20} proposed ScoreGAN with synthetic reviews generated by a Generative Adversarial Network, and the performance of a model trained on the Yelp dataset is improved. Shehnepoor \textit{et al.} thus suggested that there is a high probability of bot review presence in the Yelp dataset. A closer look at the generated review set reveals that generated reviews are often short. This is because generating long semantically coherent reviews requires a large training dataset. The syntax is not an issue, as there are only a relatively small set of syntax rules. 
A possible future direction is to distinguish between human-written fraud reviews and bot-generated fraud reviews. This will provide useful hints to characterize both human and bot-generated reviews. 
Few studies investigated such a problem and the studies mostly suffered from uncertainties in determining whether the final results can be validated unless the content is synthetically generated.   
Shehnepoor \textit{et al.} also discussed the importance of using the rating in combination with the review text, as the only behavioral indicator of bots. 

In summary, these findings suggest that bots can be identified with three key factors: \textbf{first,} certain distribution over characters, or words;  whether synthetically correct, but semantically not; and \textbf{finally,} coherence between text and rating. 

% The generator in GAN has possibly a great part to play the role in imitating such fraud generation systems, as suggested by \cite{shehnepoor20,Aghakhani2018}.  \\

% Wang \textit{et al.} \cite{wang2017handling} adopt three text-based statistical features (Review Length (RL), Rating Deviation (RD), Maximum Content Similarity (MCS)) are the only features which can be used to provide similarity measures between a new user and existing one, make it a suitable choice for handling the cold-start problem. One may use text aggregation for Multi-component classification. 
\subsubsection{Cold-start Problem}
The cold start problem can be expanded to a multi-domain (hotels, restaurants, online shops, etc.) problem, where different domains are involved in tracking the behavior of a user. A fraudster typically does not only write reviews on one platform, but multiple ones to maximize the impression. Similar studies are conducted in recent years to find the path of influential users in different social networks, namely multi-layer networks~\cite{future-work-al2016identifying,future-work-oselio2014multi}. As previously mentioned, fake news detection studies often employ a similar approach to verify the accuracy of  a news article. In a similar study, Sivasankari \textit{et al.}~\cite{future-work-sivasankari2021tracing} used a multi-domain series of data to trace back the truth about suspicious news. Such cross-domain behavior analysis with a focus on similar patterns can be useful~\cite{you2018attribute}. Although You \textit{et al.}~\cite{you2018attribute} employed a similar approach towards the cold-start problem, the study lacks the behavioral pattern exploration and analysis and solely relies on extracting attributes from multiple domains.

Graph-based approaches \cite{wang2011review} are potentially suitable for the classification/scoring step to train the joint representation and address different challenges (e.g., cold-start problem). Graph-based learning considers the relations between the extracted features to refine the prior knowledge using a network-based inference algorithm. The inference algorithms mainly embed the network-based representations of the components into a single representation. Some previous graph-based approaches (e.g., Loopy Belief Propagation ~\cite{Akoglu2013,Shebuit2015}) suffered from limitations in normalizing the nodes’ representation. The limitation leads to incorrect component predictions, since the feature aggregation is simply altered once the node’s degree is changed. This allows fraudsters to manipulate their behaviors through camouflage. One possible solution is to employ graph-based inductive learning~\cite{Hamilton2017}. Inductive learning facilitates the representation learning for single or multiple components in a graph. The representations can then be used to find the primary representations of the newly introduced nodes, such as users or items in a social platform to handle the cold-start problem.

In summary, a potentially more optimized solution is to combine 1) user path prediction over a multi-layer network and 2) using the cross-domain features (as suggested by~\cite{you2018attribute}) to find similar co-occurring behavioral patterns, or 3) graph-based approach to propagate the representation to unseen data.

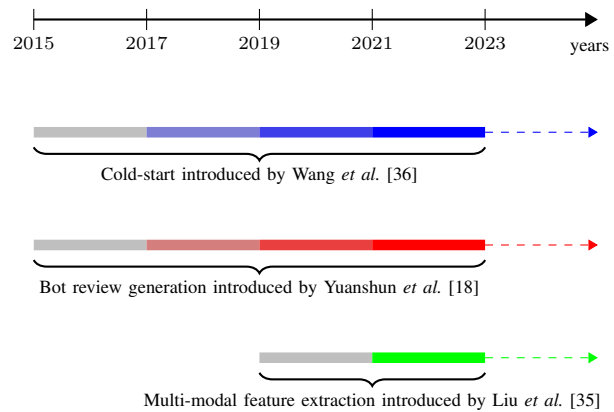
\begin{figure}
    \centering
    \begin{tikzpicture}
    % draw horizontal line   
    \draw[thick, -Triangle] (0,0) -- (7.5,0) node[font=\scriptsize,below left=5pt and -8pt]{years};
    
    % draw vertical lines
    \foreach \x in {0,1.5,...,6}
    \draw (\x cm,4pt) -- (\x cm,-4pt);
    
    \foreach \x/\descr in {0/2015,1.5/2017,3/2019,4.5/2021,6/2023}
    \node[font=\scriptsize, text height=1.75ex,
    text depth=.5ex] at (\x,-.3) {$\descr$};
    
    % colored bar for cold-start
    \foreach \x/\perccol in
    {0/100,1.5/66,3/33,4.5/0}
    \draw[lightgray!\perccol!blue, line width=4pt] 
    (\x,-1.5) -- +(1.5,0);
    \draw[-Triangle, dashed, blue] (6,-1.5) --  +(1.5,0);
    
    % colored bar for generation
    \foreach \x/\perccol in
    {0/100,1.5/66,3/33,4.5/0}
    \draw[lightgray!\perccol!red, line width=4pt] 
    (\x,-3) -- +(1.5,0);
    \draw[-Triangle, dashed, red] (6,-3) --  +(1.5,0);
    
    % colored bar for multi-modal
    \foreach \x/\perccol in 
    {3/100,4.5/0}
    \draw[lightgray!\perccol!green, line width=4pt] 
    (\x,-4.5) -- +(1.5,0);
    \draw[-Triangle, dashed, green] (6,-4.5) --  +(1.5,0);
    
    % braces
    \draw [thick ,decorate,decoration={brace,amplitude=5pt}] (6,-1.7)  -- +(-6,0)
           node [black,midway, font=\scriptsize,below=4pt] {Cold-start introduced by Wang \textit{et al.} \cite{wang2017handling}};
    \draw [thick,decorate,decoration={brace,amplitude=5pt}] (6,-3.2) -- +(-6,0)
           node [black,midway,font=\scriptsize, below=4pt] {Bot review generation introduced by Yuanshun \textit{et al.} \cite{Yuanshun2017}};
    \draw [thick,decorate,decoration={brace,amplitude=5pt}] (6,-4.7) -- +(-3,0)
           node [black,midway,font=\scriptsize, below=4pt] {Multi-modal feature extraction introduced by Liu \textit{et al.} \cite{liu2019opinion}};
    
    \end{tikzpicture}
    \caption{Emerging areas timeline and potential future work.}
    \label{fig:future-timeline}
\end{figure}

% Text is no exception and not surprisingly the fraudsters used such techniques in their favor to generate the intended contents; namely, bot-generated reviews. 
We summarized the future works in a timeline depicted in Fig.~\ref{fig:future-timeline}, highlighting the three current hot topics for fraud detection. We picked an open problem in each of the three stages in our proposed systematic framework (component selection, feature extraction, classification/scoring) and elaborated on the current study. Bot-generated review detection was introduced as an open problem in the component selection step by Yuanshun \textit{et al.}~\cite{Yuanshun2017} and was primarily investigated by two studies by Aghakhani \textit{et al.}~\cite{Aghakhani2018} and Shehnepoor \textit{et al.}~\cite{shehnepoor20}. As explained in the feature extraction step, the main challenge in the feature extraction step is to find a suitable representation for the selected component. Multi-modal data can effectively address such a challenge, as introduced by Liu \textit{et al.}~\cite{liu2019opinion}. Extracting data from different network layers of the OSI model~\cite{Wang2019FdGars,Wen2020} results in better activity modeling of reviewers in a social review platform, and thus provides a more informative feature representation. Graph-based approaches can be treated as a scoring method to address different challenges such as the cold-start problem (introduced by Wang \textit{et al.}~\cite{wang2017handling}) through its capability in modeling complex relationships between the components.

% {\color{blue}
% \subsubsection{Findings}
% % We discussed the significance of the labelled data and the expertise from other areas in improving the quality of the labelling in Sec.~\ref{sec:data}. To overcome the possible data scarcity problem, assuming the lack of labelled data is still at the place, we mostly discussed transfer learning in Sec.~\ref{sec:topic-anal}. With new topics arising in fraud detection, there could be a potential to apply the suggested findings to new topics introduced in Sec.~\ref{sec:future-works}.\\
% \textbf{First,} 
% \textbf{The next challenging topic} is bot review detection. 
% }

\section{Conclusion}
\label{sec:conclusion}
With an increase in the popularity of social review platforms for businesses and their customers, the traditional marketing media (e.g., broadcast and print) are losing ground to such web and mobile-based social review platforms. Fraudsters seize such opportunities for their benefits and take advantage of the media to write fraud reviews on targeted businesses. Characterizing fraud reviews provides an opportunity for new researchers to better understand the nature of such reviews, and hence improve the detection algorithms' performance. Several studies provided an overview to define fraud and examined the detection algorithms from different perspectives~\cite{heydari2015detection,crawford2015survey,vidanagama2020deceptive}. However, researchers may still be confused about the overall fraud detection framework. In this study, we proposed a systematic framework for fraud review detection with three stages: component selection, feature extraction and representation, and classification/scoring.  
% The drawbacks of such misconducting content are not only limited to the users but also the businesses who provide the products and services since users' feedback plays an incredible role in their decisions. With recent advances in Machine Learning techniques, the impact of rotten content is both more destructive and largely spreading in the social review platform. So detecting such contents and blocking their spread has been a concern of many researchers. 
% Studies in fraud detection can be seen from different angles. One may focus on the fields of study such as detection or generation. 
\begin{itemize}
    \item In the component selection step, all studies considered either review, user, or item as their selected component to explore. While primary studies focused on fraud review detection, user group detection has recently attracted the attention of researchers, but it is still far from being fully resolved. Bot generation is another challenging topic due to the capability of deep learning-based review generation. Targeted item detection is also a potential topic to consider, as it is not investigated as a separate component and has always been studied alongside user and review for multi-component classification. 
    \item We analyzed feature extraction across three different categories; text-based, behavioral, and joint features. With recent progress in deep learning, new word embedding techniques have been employed to extract features and address the limitations of lingo-statistic text-based features. Behavioral features were also investigated individually or in a combination with text-based features as joint features. 
    \item Classification/Scoring techniques were categorized based on the approaches used to deal with the lack of reliable data. Semi-supervised and unsupervised learning deals with partial labeling data for detection, while deep learning can suggest a considerable classification improvement over classic approaches. Data augmentation approaches such as GANs are a possible solution to address data paucity. 
\end{itemize}
% Due to the lack of reliable data classification/scoring techniques were divided based on the approaches. 
In addition to the technical challenges, new problems such as cold-start, bot fraud generation, and user group detection are becoming increasingly important. One possible direction for user group detection is to use graph-based solutions (e.g., Graph Convolutional Network) due to their effectiveness in realizing the pair-wise tightness of members of a group. Multi-modal data can also be considered as an important future work since it provides a comprehensive representation of all the different OSI model layers associated with an online social review platform to describe a component.
To conclude, in this study, we introduced a systematic framework for fraud detection tasks. The techniques outlined for each step showed improvements in fraud detection. It is hoped that this review will encourage researchers to deepen their understanding of fraud detection and provide solutions to emerging challenges.
% The techniques for each step demonstrated the improvement in fraud detection. 
% Hopefully, the current review study encourages the researchers to better understand the fraud detection task and provide solutions to some of the upcoming challenges.

\bibliographystyle{IEEEtran}
\bibliography{references}

% \appendix









\end{document}